\documentclass[10pt,submit]{aiaa-tc} % insert '[draft]' option to show overfull boxes, handcarry/submit
\usepackage{epsf,exscale,times}
\usepackage{varioref}
\usepackage{ams}
\usepackage{graphicx}
\usepackage[bf]{subfigure}
\usepackage{epsfig}
\usepackage{wrapfig}
\usepackage{lettrine}
\usepackage{amssymb}
\usepackage{amsthm}

\usepackage{booktabs}
\usepackage{setspace}

\usepackage{amsmath}

\usepackage{tikz}
 \usetikzlibrary{patterns}        
 \usetikzlibrary{arrows,automata}
\usetikzlibrary{calc}

 \title{Adversary-Agent Reinforcement Learning 
  for Pursuit--Evasion} % based on StarCraft}
 
 \author{Xun Huang %
    \thanks{Professor. State Key Laboratory of Turbulence and Complex Systems, Department of Aeronautics and Astronautics. Also as Adjunct Professor at the Department of Mechanical and Aerospace Engineering, The Hong Kong University of Science and Technology. AIAA Associate Fellow. huangxun@pku.edu.cn.  }
\\   
 {  \normalsize\itshape
   College of Engineering, Peking University, Beijing, 100871, China PRC}\\
 }

\date{}
\begin{document}

\maketitle

\begin{abstract}
A reinforcement learning environment with adversary agents is proposed in this work for pursuit--evasion game in the presence of fog of war, which is of both scientific significance and practical importance in aerospace applications. One of the most popular learning environments, StarCraft, is adopted here and the associated mini-games are analyzed to identify the potential limitations for training adversary agents. 
 The key contribution includes the analysis of the potential performance of an agent by incorporating control and differential game theory into the specific reinforcement learning environment, and the development of an adversary-agents challenge (SAAC) environment by extending the current StarCraft mini-games. The subsequent study showcases the use of this learning environment and the effectiveness of an adversary agent for evaders. Overall, the proposed SAAC environment should benefit pursuit--evasion studies with rapidly-emerging reinforcement learning technologies. Last but not least, the corresponding code is available at GitHub. 

\end{abstract}

\section*{Nomenclature}
\begin{tabbing}
  XXX \= \kill% this line sets tab stop
      $a$, $c$     \> \hspace{2mm} = \hspace{2mm}  discretization of game domain  \\       	
      $h$     \> \hspace{2mm} = \hspace{2mm}  evading trajectory \\       
      $H$     \> \hspace{2mm} = \hspace{2mm}  unit health \\       
       $l_{x, y}$     \> \hspace{2mm} = \hspace{2mm}  the side length of game domain  \\       
       $J$     \> \hspace{2mm} = \hspace{2mm} linear quadratic cost \\
        $K$     \> \hspace{2mm} = \hspace{2mm}  searching steps \\
       $M$     \> \hspace{2mm} = \hspace{2mm}  discretized blocks of game domain \\
       $N_{e, p}$     \> \hspace{2mm} = \hspace{2mm}  the number of evaders/pursuers \\
       $\mathfrak{N}$     \> \hspace{2mm} = \hspace{2mm}  the number of the captured evaders  \\     
      $p$     \> \hspace{2mm} = \hspace{2mm}    the probability of capture \\                 
      $r$     \> \hspace{2mm} = \hspace{2mm}   the discovery or attacking radius \\   
           $Q$     \> \hspace{2mm} = \hspace{2mm}   game domain \\          
      $R$     \> \hspace{2mm} = \hspace{2mm}   the longest length inside game domain \\   
      $s$     \> \hspace{2mm} = \hspace{2mm}   searching trajectory \\   
       $\mathfrak{R}$     \> \hspace{2mm} = \hspace{2mm}  reward \\   
      ${T_k}$     \> \hspace{2mm} = \hspace{2mm}  defeated time of all evaders \\          
            ${T_f}$     \> \hspace{2mm} = \hspace{2mm}   game duration \\       
      $U$     \> \hspace{2mm} = \hspace{2mm}   the moving speed of units \\       
      $v$     \> \hspace{2mm} = \hspace{2mm}   the expected capture time  \\        
       $x$, $y$     \> \hspace{2mm} = \hspace{2mm}  Cartesian axis  \\       	
      $\mu$     \> \hspace{2mm} = \hspace{2mm}  the area of game domain \\    
     % $\delta$     \> \hspace{2mm} = \hspace{2mm}  a small value close to 0 \\        	

  \textit{Superscripts}\\
	 $(\cdot)^{*}$    \> \hspace{2mm} = \hspace{2mm} the optimal solution of ($\cdot$) \\
	
  \textit{Subscripts}\\

 %$(\cdot)_{0}$    \> \hspace{2mm} = \hspace{2mm} stationary value of ($\cdot$) \\
 $(\cdot)_{e, p}$    \> \hspace{2mm} = \hspace{2mm} variables related to evaders/pursuers \\

 \end{tabbing}

\section{Introduction}
A reinforcement learning environment is developed in this paper for pursuit--evasion game,  which is a classical but challenging problem with important aerospace applications, such as simultaneous and cooperative interceptions \cite{Pachter:19AIS,Zadka:20jgcd} and exoatmospheric interception \cite{Shen:18jgcd,Gutman:19jgcd, Ye:20AST, Venigalla:21jgcd} and search-and-rescue operations \cite{Leone:21}. The problem has been studied extensively under the analytical framework of differential game theory \cite{Gal:79SIAM,Sun:21siam} and optimal control theory \cite{Gutman:19jgcd,Car:18jgcd}, respectively. Recently, the merging of game theory, control theory and deep learning has become a popular topic \cite{Cao:20nsr}. Wang \textit{et al.} have proposed a distributed cooperative pursuit strategy based on reinforcement learning  and performed tests in the openAI  Predator--Prey environment \cite{Wang:20neur}. Multi-agent reinforcement learning has been further considered for pursuit--evasion with multiple unmanned aerial vehicles \cite{Wang:20ACA}.  Li \textit{et al.} have proposed an estimation algorithm of optimal pursuing strategy based on Thompson sampling \cite{Li:18PE} and conducted tests in Atari Pac-Man environment. Most of those pioneering works have essentially focused on artificial intelligent (AI) strategies only for one player (usually the pursuer), while the other player is either immobile or cannot be directly controlled by another AI agent, which actually reduces possible conflicting levels of pursuit--evasion game. To address this issue, the current work endeavors to develop a reinforcement learning environment for pursuit--evasion, both of which can then be directly controlled by a separate agent, through the extension of the famous StarCraft II game environment.

StarCraft II is one of the most popular real-time strategy games currently played worldwide. The game requires human players making very rapid decisions on strategical, tactical, and economical levels. To study AI's capability, DeepMind has recently developed a Python interface library, PySC2 \cite{PySC2}, which exposes StarCraft II's low-level application programming interface to a reinforcement learning environment. The combination of PySC2 and StarCraft II learning environment has enabled deep learning studies of competition and coordination within multiple agents \cite{Samvelyan:19smac}  in complex environment with representative terrains (cliff, water, forest, etc.) and  partial observation (so-called fog of war \footnote{In the game, fog of war means that a player cannot observe the information of the region of map when the region is not close to the player's units, buildings, or scouting abilities. It is represented by dark region on both the radar map and the main screen.}). An agent from DeepMind, AlphaStar \cite{Arulkumaran:AlphaStar}, has achieved grandmaster level performance by beating top professional players \cite{Vinyals:19nature}. Nevertheless, it is worthwhile to mention that the training of AlphaStar requires thousands of GPU processors and the cost was estimated to be more than 10 million USD \cite{alphastar_cost}, which is prohibitive for small research groups.

To reduce complexities from full StarCraft games, DeepMind has further provided seven single-player, fixed length  mini-games to explore deep learning capabilities on various specific tasks. In which, the so-called FindAndDefeatZerglings mini-game is very similar to one type of classical pursuit--evasion games in dark room (also known as the princess and monster game \cite{Gal:79SIAM}). However, the author argues that the following unsolved issues handicap the existing StarCraft  II mini-games 
to be a viable deep learning environment for pursuit--evasion. 
\begin{itemize}
\item First and foremost, the reason about why certain performance could be achieved by an AI agent is still unknown. When the opponent evasion units are controlled  by the low-level build-in StarCraft II code, 
the best mean score of the reward \footnote{It was defined in the mini-game as the number of the evasion units that have been found and defeated in $180\,$s.} is 46 (from Fig.~6 of the reference \cite{Vinyals:17star}, although DeepMind later mentioned this value as 61 in Table 1 of another article \cite{Zambaldi:19relational}). The same problem has been repeated by some other groups and the achieved best mean score is from 16 (with an 
asynchronous advantage actor-critic (A3C) agent \cite{Alghanem:A3C18}), 22.1 (with an  advantage actor-critic (A2C) agent \cite{reaver})  to 45 (with an A2C agent \cite{pysc2deep}). The best score achieved now is 62 through a relational agent from DeepMind \cite{Zambaldi:19relational}. An intriguing open question is that what is the best performance that could be achieved by a well-designed agent, and why? 

\item Second, all those mini-game maps from DeepMind only support single agent. The famous 
StarCraft multi-agent challenge (SMAC) toolkit in the reference \cite{Samvelyan:19smac} has provided a multi-agent reinforcement learning environment. Nevertheless, the SMAC environment also only supports to control all pursuer units with independent agents, whereas the opponent evasion units are still controlled by the built-in StarCraft II scripts and cannot be controlled by any external agent. Is it possible to develop a mini-game based StarCraft learning environment where adversary players can be controlled by opponent AI agents? 
\end{itemize}
\begin{figure}[htbp!]
  \begin{center}
%            \subfigure[]{
 % \includegraphics[height=5cm]{figures/MoveToBeacon1.jpg}}
  %            \subfigure[]{
 % \includegraphics[height=5cm]{figures/CollectMineralShards1.jpg}}
  %             \subfigure[]{
  \includegraphics[height=5cm]{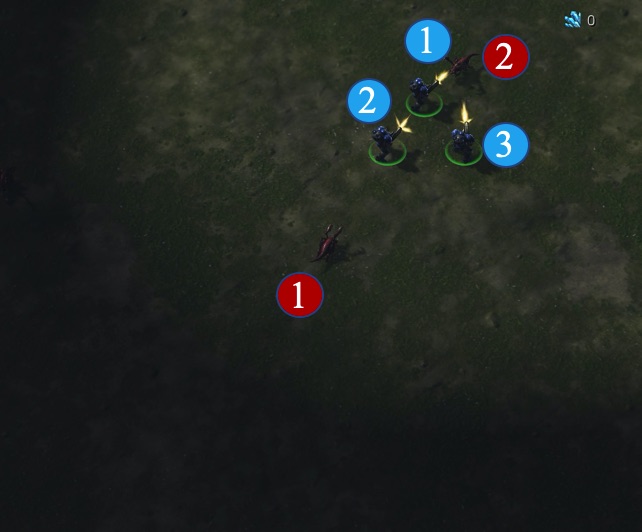} %} 
  \caption{ Screenshot of the FindAndDefeatZerglings mini-game from DeepMind, where 3 pursuers (marines, denoted by cyan circles) are searching and firing at 25 zerglings, while only a couple of them (denoted by red circles) are visible to the marines when fog of war (i.e., the dark area) is activated.  As shown below, some of these army units are replaced with aerospace units in this work to imitate pursuit--evasion game.  }
  \label{f:FindAndDefeatZerglings}
  \end{center}
\end{figure}
The above two issues directly motivate the current work. The first issue is addressed here by merging control theory and differential game theory. Isaacs has pioneered this direction by proposing theoretical strategies under a dynamic game framework \cite{Isaacs65}. Some more recent developments, especially from the numerical direction, can be found in the monograph \cite{Bernhard09}. Moreover, Gal has provided a theoretical solution  
of princess and monster game in a generic geometrical domain \cite{Gal:79SIAM}. The differences (and similarities) between Gal's problem and the FindAndDefeatZerglings mini-game (see Fig.~\ref{f:FindAndDefeatZerglings}) are highlighted in this paper, which further enables the theoretical developments that explain the achievable game score under different representative scenarios.

 The second issue is resolved by first identifying the pending programming issues inside the PySC2 interface \footnote{The most updated version 3.0 was used during the preparation of this article.} when two adversary agents are implemented. Next, some of the necessary rectifications of the corresponding code are conducted to fix the identified issues. The associated programming tricks and modifications are given to enable interested readers to utilize the learning environment and set up their own differential game problems in the future. Moreover, the source code and the extended mini-game maps that support two adversary-agents are developed in this work and available to the public at GitHub\footnote{https://github.com/xunger99/SAAC-StarCraft-Adversary-Agent-Challenge.}.  Overall, the contribution of this work is twofold: (1) to enhance the understanding of reinforcement learning capability for pursuit--evasion game by merging with control and game theories, and (2) to propose an adversary-agent reinforcement learning environment for pursuit--evasion game with progressively complicated set-ups of practical importance. 

The remaining part of this paper is organized as follows.  Section~\ref{s:SC2} will introduce the fundamentals of StarCraft II learning environment, with the focus on pursuit--evasion type game. A couple of pending unsolved issues will be highlighted therein. Then, a theoretical study will be given in Sec.~\ref{s:merge} to explain the possible game performance that can be achieved for the current pursuit--evasion game set-up. Next, Sec.~\ref{s:SAAC} will introduce the proposed adversary-agent learning environment and discuss the corresponding results, especially from the perspective of game theory. Finally, Sec.~\ref{s:cons} will summarize the present work. More background information regarding the units and code structures can be found in the appendix.

\section{StarCraft II Learning Environment}\label{s:SC2}
%[

StarCraft II is a popular and challenging real-time strategy game developed by Blizzard Entertainment. 
The basics of StarCraft II is introduced here for the completeness of this paper. For more information, interested readers can download and play this game for free.  Moreover, some important programming tricks are summarized in this section (and the appendix) for the better using of this learning environment.

The full game has a science fiction setup with three different races: Terran (supposedly to imitate human army), Zerg (mimics worm army) and Protoss (mimics alien high-intelligence army). Each race consists of a number of distinctive units with unique strength and weakness. Some of those used in this work are summarized in the appendix. The game starts by choosing a race with a number of units and resource, following by macromanagement to develop economics and build up units (many of which are aerospace units) and split-second decisions on tactical level to beat opponents, which could be computer bots, human players or intelligent agents. Due to its complexity and extremely large action space, StarCraft II has been regarded as a new challenge for reinforcement learning \cite{Vinyals:17star} after the game of Go.

\begin{figure}[htbp!]
  \begin{center}
    \includegraphics[width=12cm]{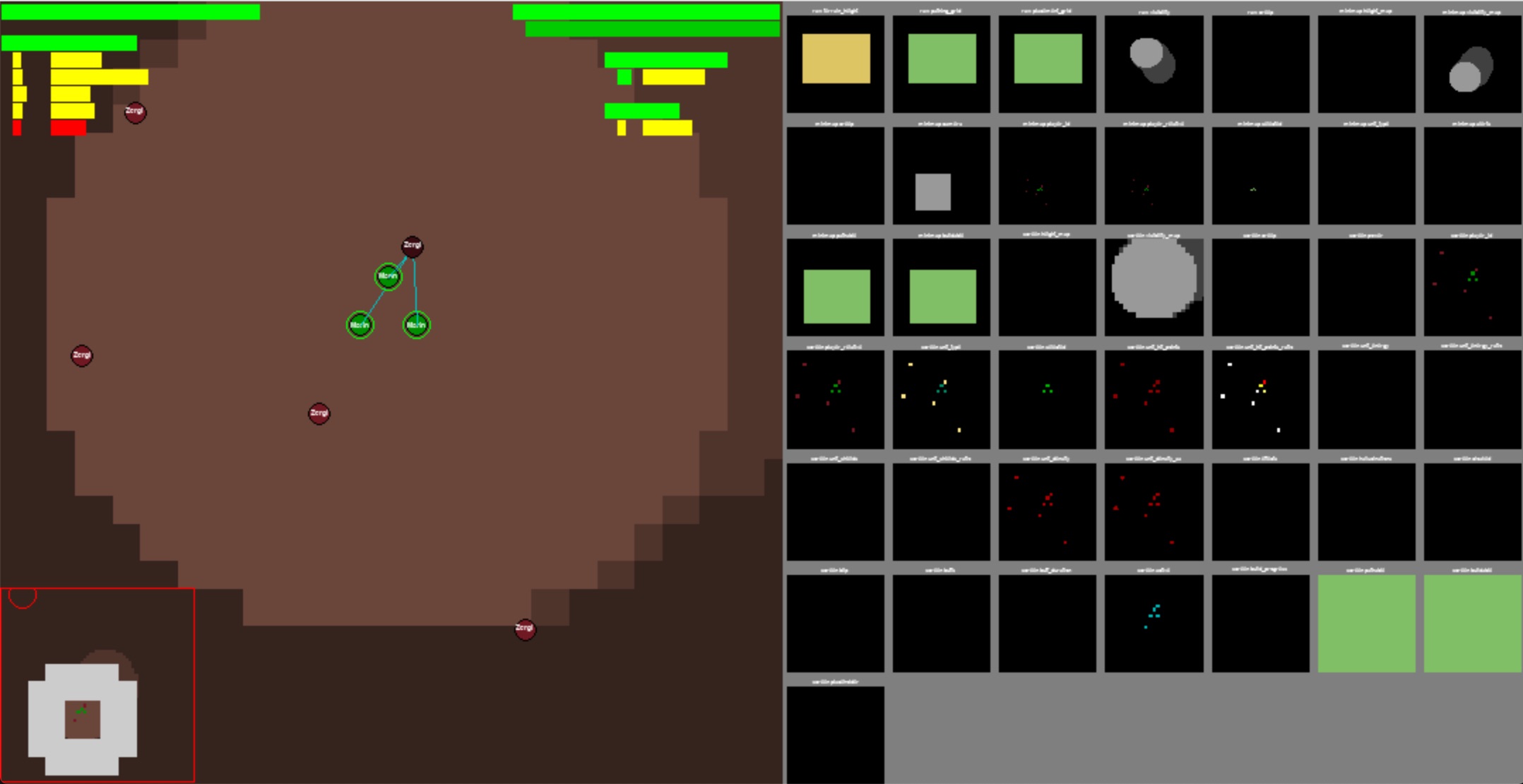}
    \caption{The feature map (left) and a number of feature layers (on the right) of height, fog of war, camera locations, alien and opponent units, etc., can be obtained instantaneously from PySC2 when the StarCraft II game is played by an agent.  }
    \label{f:features}
  \end{center}
\end{figure}

Recently, DeepMind has developed a Python interface library, PySC2 \cite{PySC2}, which enable users to obtain spatial observations (in a form of features, see Fig.~\ref{f:features}) and to learn to conduct humanlike actions. In addition, several game score/rewards can be accessed to examine how well an AI agent is working. Users shall design an appropriate 
score for their own learning tasks to differentiate the performance of  agents.
As an example, Fig.~\ref{f:FindAndDefeatZerglings} shows an instantaneous screenshot of the FindAndDefeatZerglings mini-game, where three marine units (from Terran) should be trained to explore the  two-dimensional (2D) domain \footnote{The domain can be easily extended to three-dimensional by including different terrain elements in the StarCraft map editor.} activated with fog of war to find and defeat 25 individual zerglings (from Zerg) that have been randomly deployed throughout the map. More information of each units, regarding health, detecting range and attacking range, etc., can be found in the appendix. 

%The mini-game FindAndDefeatZerglings is unique in all mini-games for two reasons: (1) the dark area is unknown to the three marine units, and (2) the training cost is particularly large.  (xxxx by code TestScripted\_V1.py)
%
%\begin{itemize}
%\item MoveToBeacon, where a marine unit is trained to move to a randomly located beacon; 
%\item CollectMineralShards, where two marine units are trained to rapidly move and collect mineral shards spread around the map; 
%\item FindAndDefeatZerglings, where three marine units are trained to explore a map with fog of war to find and defeat individual zerglings.
%\end{itemize}
% of war means that a play
\iffalse
\begin{figure}[htbp!]
  \begin{center}
            \subfigure[]{
  \includegraphics[height=5cm]{figures/MoveToBeacon1.jpg}}
              \subfigure[]{
  \includegraphics[height=5cm]{figures/CollectMineralShards1.jpg}}
               \subfigure[]{
  \includegraphics[height=5cm]{figures/FindAndDefeatZerglings1.jpg}} 
  \caption{ The three mini-games: (a) MoveToBeacon, (b) CollectMineralShards, and (c) FindAndDefeatZerglings. }
  \label{f:mini-games}
  \end{center}
\end{figure}
%
\fi

\begin{figure}[htbp!]
  \begin{center}
            \subfigure[]{
  \includegraphics[width=5cm]{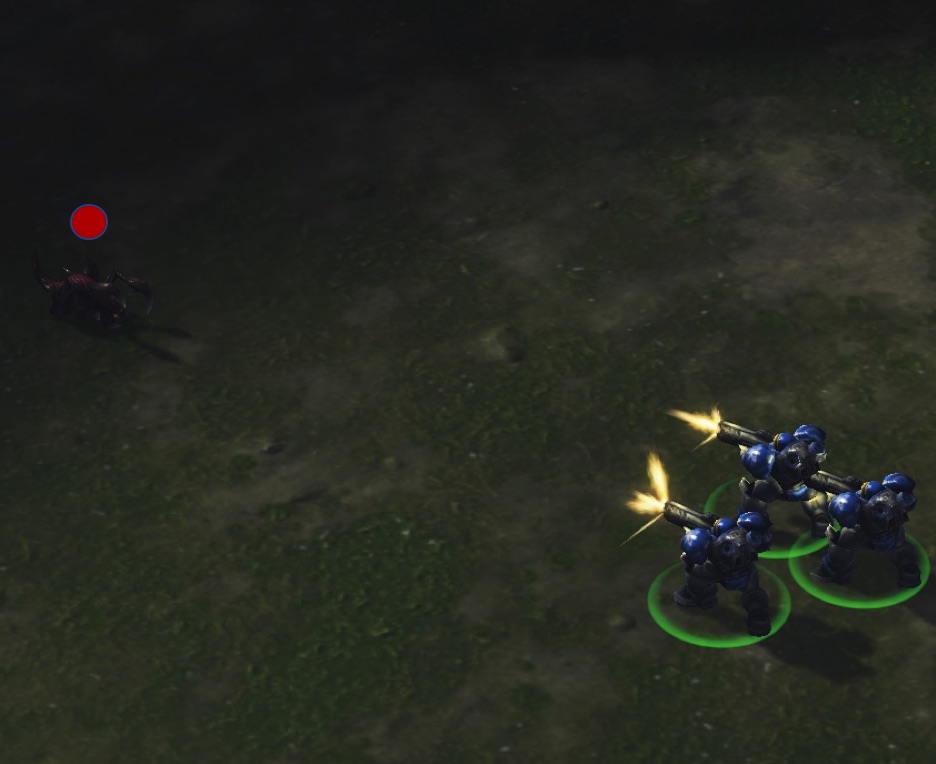}}
              \subfigure[]{
  \includegraphics[width=5cm]{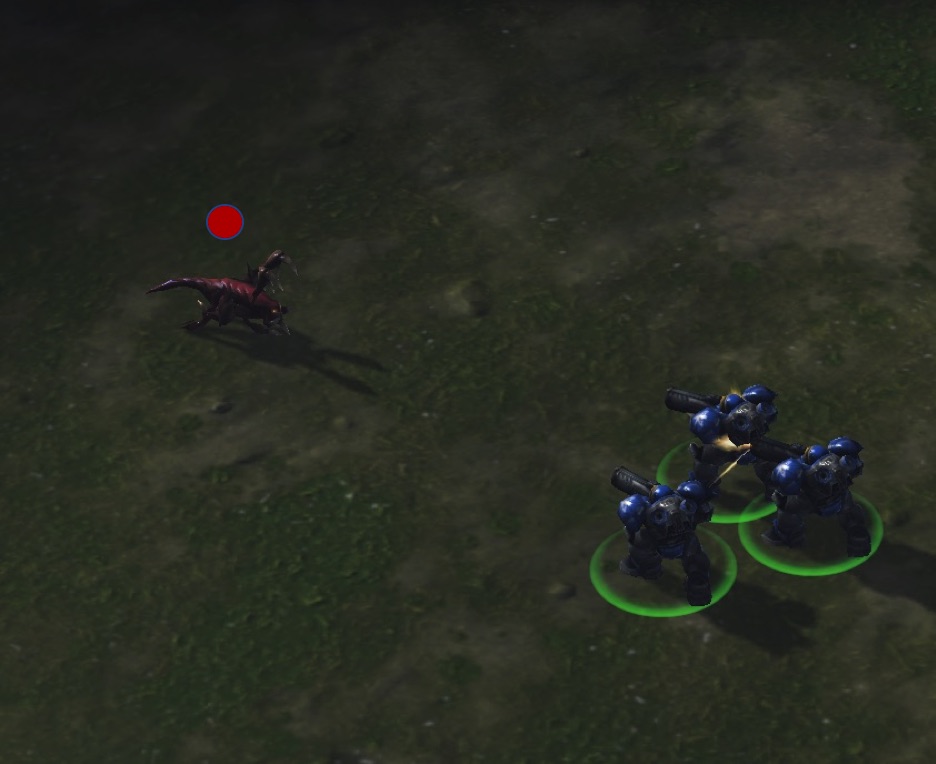}}
            \subfigure[]{
  \includegraphics[width=5cm]{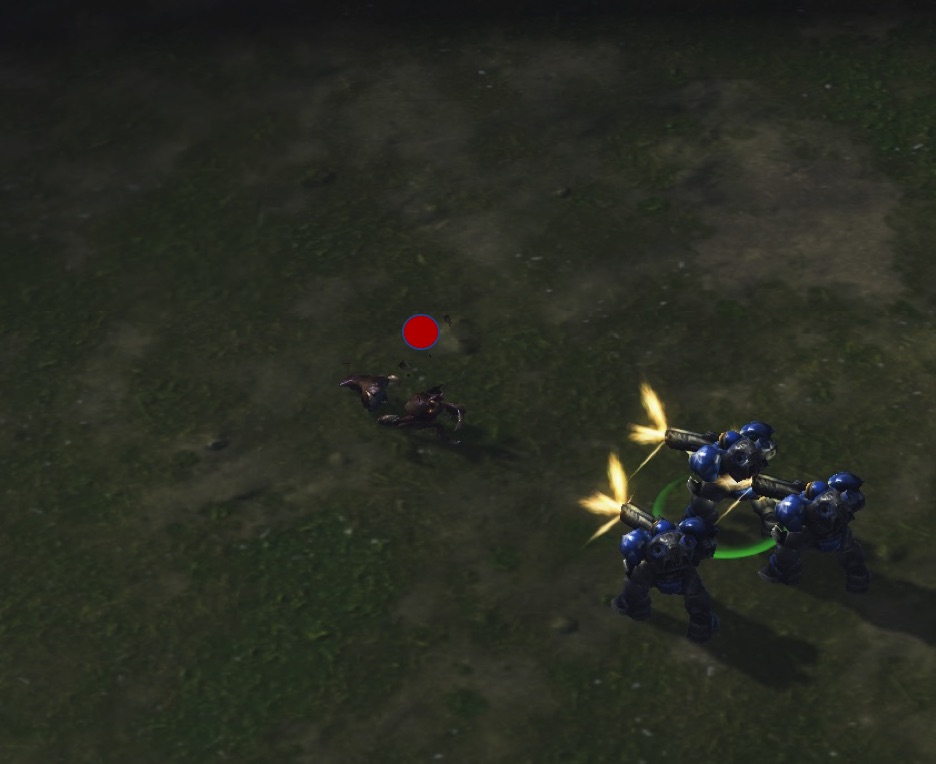}}
  \caption{ In the mini-game of FindAndDefeatZerglings: (a) the three marines will search and attack the zergling (denoted by the red spot in each panel) at the verge of the fog of war (the dark area); (b) the zergling will run towards marines and try to push back, instead of escaping into the fog of war nearby; and (c) the counterattack will be easily defeated by the concentrated fire from the three marines.    }
  \label{f:counterattack}
  \end{center}
\end{figure}
\begin{figure}[htbp!]
  \hspace{25mm}
            \subfigure[]{
  \includegraphics[width=3.5cm]{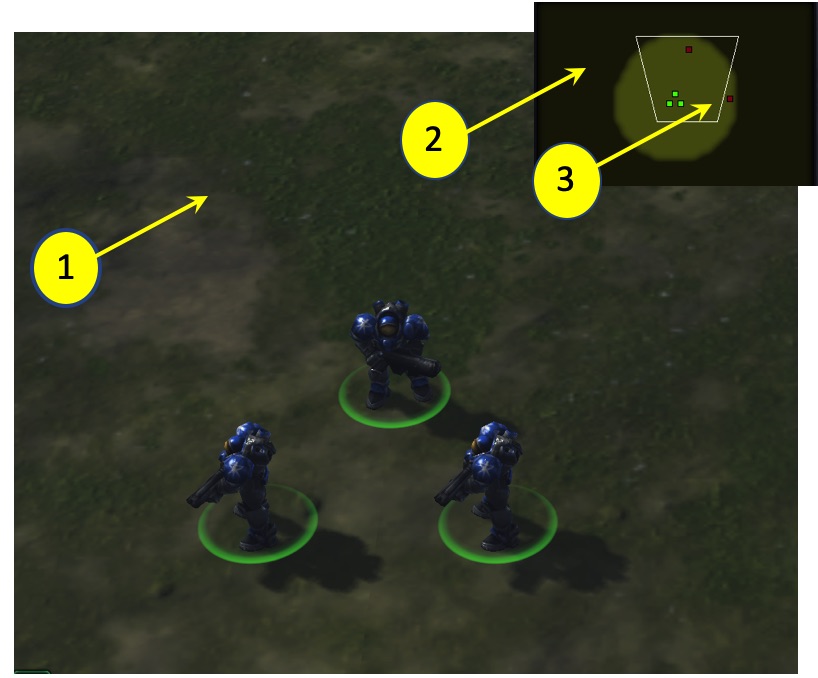}}
  \hspace{3mm}
              \subfigure[]{
  \includegraphics[width=3.5cm]{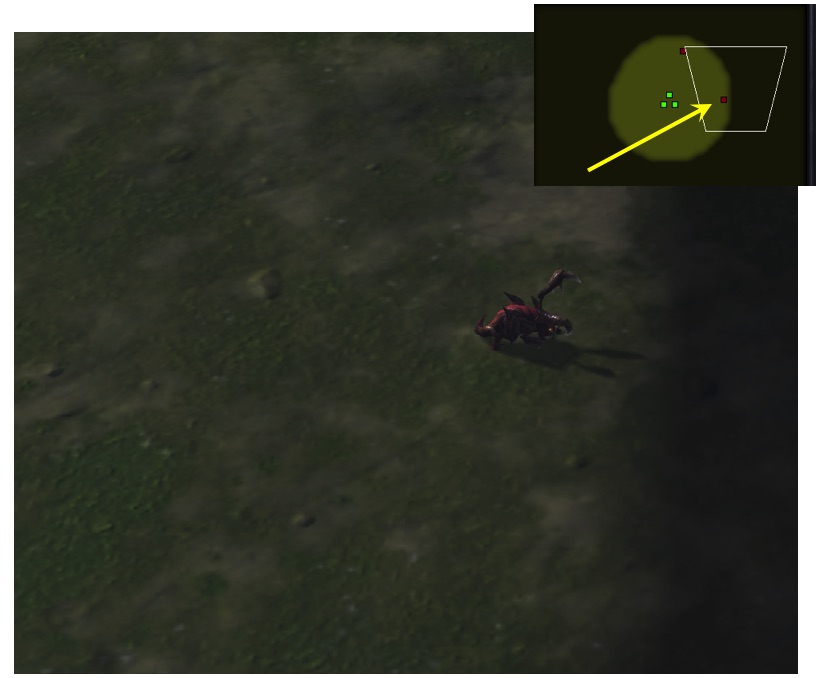}}
    \hspace{3mm}
            \subfigure[]{
  \includegraphics[width=3.5cm]{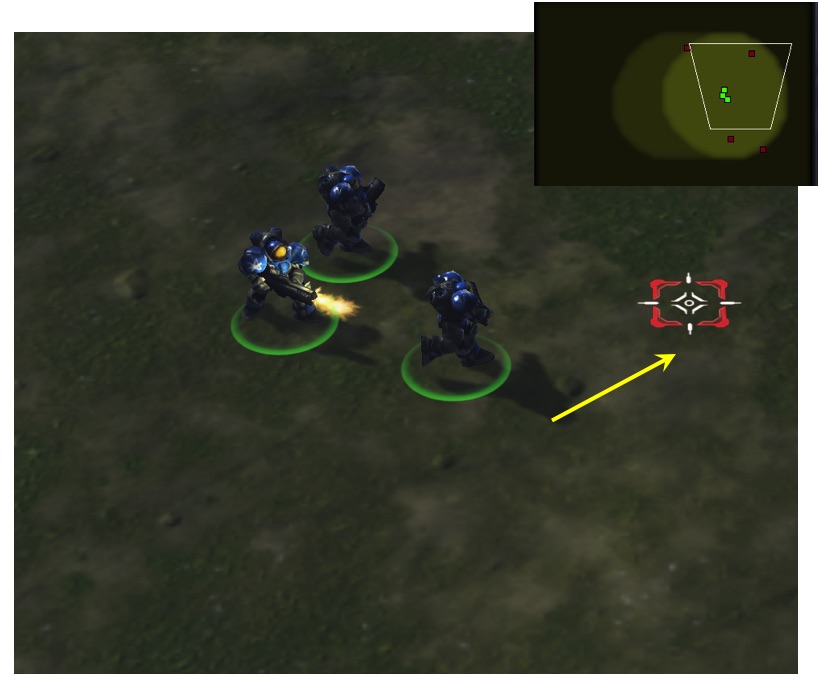}}
    \begin{center}
  \begin{subfigure}[]{
\begin{tikzpicture}
	\draw [thick, gray] (-9,0) -- (7,0);
	\draw [thick, gray] (7,-3) -- (-9,-3);
	\draw [thick, gray] (-9,-1.5) -- (7,-1.5);
	\node [above] at (-8.,-1) {Human actions:};
	\node [above] at (-8.,-2.5) {Agent actions:};
	\node [above] at (-5,-1) {Select all units};
	\node [above] at (-5,-2.5) {select\_army};
	
	\node [above] at (-1,-1) {Move camera};
	\node [above] at (-1,-2.5) {move\_camera};

	\node [above] at (3,-1) {Attack};
	\node [above] at (3,-2.5) {Attack\_screen};
	
	\node [above] at (6,-1) {IDLE};
	\node [above] at (6,-2.5) {no\_op};
\end{tikzpicture}}
\end{subfigure}
  \caption{ When the StarCraft learning environment is adopted, the agent actions should be coded by following the action habits (and frequency) of human players. For example: (a) a human player will select the marines under the present camera view (1), then observe the radar (2) to search zerglings, and could find one (3) to the right of the radar; (b) next, the human player will move the camera to the founded zergling (with click of mouse), and (c) then issue the attack action (with the shortcut key  \lq a\rq~ plus click of  mouse).  An agent should be designed to follow the same action steps, which are summarized in (d). Oterhwise, the StarCraft environment will return unanticipated action results.   }
  \label{f:actions}
  \end{center}
\end{figure}

Immediately, it can be seen \footnote{Install StarCraft and PySC2, then type the terminal command: python -m pysc2.bin.agent --map FindAndDefeatZerglings.} that the mini-game FindAndDefeatZerglings is quite similar to pursuit--evasion game, with the loss of the evaders being the number of the defeated zerglings until the game is finished (at $T_f=180\,$s). Nevertheless, compared to game set-ups in most former theoretical works, some distinctive differences of the current game set-up and the associated effects on pursuit--evasion can be identified as follows.
\begin{enumerate}
\item The mini-game contains fog of war, which extensively increases the game complexity to such a level that the corresponding learning speed is much slower than any other mini-games with deactivated fog of war. 
\item A pursuit--evasion game is a two-person zero-sum game, mostly only consisting of 1 pursuer and 1 evader.  More complicated set-ups have been considered in the recent work \cite{Zadka:20jgcd}, where several (ground) evaders are protected by many defending (airborne) pursuers that call for simultaneous attacking strategies. Similarly, the FindAndDefeatZerglings mini-game consists of 3 pursers and 25 evaders.    
\item  In the FindAndDefeatZerglings mini-game, instead of running away, the build-in code from StarCraft will control the evaders (i.e., zerglings) run towards and attack the pursuers (i.e., marines) when both are within the sight range (see Fig.~\ref{f:counterattack}). Hence,  the mini-game is not a typical pursuit--evasion game. However, as shown in the appendix, the attacking capability of a couple of evaders is much worse than the three pursuers. Hence, the attacking (or self-defense) action from the evaders will actually simplify the searching/exploration tasks of the pursuers. As to be shown below, such an action is deliberately disabled and only the evasion action would be allowed in the proposed new adversary learning environment. 
\item The mini-game from DeepMind only supports one agent to control the three marine units. The other build-in low-level code controls the evaders, which will either remain still when no opponent is within the sight range or otherwise rush towards and attack the pursuers. Hence, the current FindAndDefeatZerglings mini-game imitates pursuit--evasion game with immobile evaders. 
\item Last but not least, in StarCraft/PySC learning environment, a deep learning design is supposed to mimic (and then rival) the intelligence of human players.  Figure~\ref{f:actions} gives an example. It is natural to 
expect that an AI agent should directly pursuit any evaders that have been found on the radar. A human play, however, must first move the camera view to the target area and then issues the  pursuing action. Hence, when the StarCraft II environment is used, all agent actions must be designed to follow the operation/behavior habit of a human player.   
\end{enumerate}
\noindent The above issues 1, 2 and 5 of the StarCraft learning environment increase the problem complexity of pursuit--evasion game. The issues 3 and 4 are resolved in the proposed new learning 
 environment and details can be found in Sec.~\ref{s:SAAC}.

It is worthwhile to mention that many other popular reinforcement learning environments can be found, e.g.,  from the famous toolkit of Gym and Atari environment \cite{gym, Hasselt:15DQN}. Compared to those reinforcement learning game problems, the StarCraft II learning environment is challenging because of the large size of available action space. For example, the size of the action space for the classical inverted pendulum control problem from Gym is 2 (i.e., either push left or right). The size for the Pac-Man game is 4 (i.e., go top, bottom, left or right). In contrast, the size of the action space for StarCraft II learning environment is around 300, including move, attack, management tasks, etc., further with 13 types of possible arguments.  One may argue that most of these StarCraft II actions are unnecessary for  
pursuit---evasion game and can be easily disabled from the available action space during the reinforcement learning of an agent. Then, it seems that the size of the action space can be reduced to 4  
(refer to agent actions in Fig.~\ref{f:actions}). However, 
the actual pursuing or evasion coordinates, whose size is $32 \times 32$ for the current mini-game set-up, must be given along with the associated action command. Generally speaking, the spatial coordinate outputs shall be regarded as a part of the action space.  
Then, for the current pursuit--evasion type mini-game, the size of the action space is  $32 \times 32$+ 4 = 1028. For such a complicated set-up, it should be beneficial to first have a theoretical study from the perspective of control and differential game theory. 

 %As a result, the deep learning speed for the pursuit--evasion mini-game is much slower than other mini-games.  
%

\iffalse
\begin{itemize}
\item select: select units to under control; 
\item move: move the selected units to target coordinates;   
\item attack: attack the targeted coordinates; 
\item camera: control the camera;  
\item no\_op: do nothing, actually equal to continue; 
\item coordinates: the spatial parameters of the actions. 
\end{itemize}
\fi
%
%

%The problem can be classified based on the number of players: one/multiple pursuer(s) versus one/multiple evader(s) games, where the adversaries' potential strategies are inputs to the synthesis of each actions. 

\section{Merging with Control and Game Theory}\label{s:merge}
\subsection{A linear control perspective}

The pursuit--evasion game has a long-lasting connection to control theory. Normally, a strategy of the  game consists of two levels, where the bottom level is control level, which is usually the well-known proportional guidance law, while the top level is the pursuing or evading strategy  \cite{Shneydor98}. Moreover, the control perspective helps to show that why the normal linear optimal control designs must merge with deep learning for pursuit--evasion game especially when fog of war is activated. 

For the current pursuit--evasion set-up, the state dynamics is absent \footnote{Acceleration capability is neglected here, though some units inside StarCraft, such as marines and medivac dropship, do have the capability, which could be considered in future studies.} and the state space representation is simply 
\begin{eqnarray}
\dot{\mathbf{x}}_e &=& \mathbf{u}_e, \\ 
\dot{\mathbf{x}}_p &=& \mathbf{u}_p,  
\end{eqnarray}
\noindent where $\mathbf{x}$ is the 2D coordinates (states) of all the units, $\dot{(~~)} \triangleq d/dt$, $\mathbf{u}$ is the moving velocity (control inputs), and the subscripts $(~~)_p$ and $(~~)_e$ represent the pursuers and evaders, respectively. 
A more generic game set-up with nonzero state dynamics and the corresponding theoretical manipulation can be found in the reference 
\cite{Sun:21siam}. 

Following \cite{Gutman:19jgcd}, a linear quadratic cost can be defined for each pair of the pursuer and evader (e.g., $\mathbf{x}_{p_i}$ and $\mathbf{x}_{e_j}$) with a specified final time $T_f$ (i.e., $180\,$s in the current mini-game set-up),     
 \begin{eqnarray}
J_{ij} = || \mathbf{x}_{p_i} - \mathbf{x}_{e_j} ||^2 + R_p \int_0^{T_f} || \mathbf{u}_{p_i} ||^2 dt,   
\end{eqnarray}
 \noindent where the weight $R_p$ can be relaxed to 0 since there is no penalty on control input.  
 
In the mini-game, the number of the defeated units $\mathfrak{N}$ is defined as the loss of the evaders and the gain of the pursuers, that is, when $J_{ij}$ is minimized to 0 (or just within the attacking range of pursuers), the $i\,$th pursuer will be able to attack the $j\,$th evader. 
%the pursuit--evasion game requests to capture as many of evaders as possible, that is,  
 %
%  \begin{eqnarray}\label{e:num}
%\mathfrak{N}= \mathrm{max}\left( \mathrm{min} || (J_{ij}) ||_0 \right),   ???
%\end{eqnarray}
 %
 %\noindent where $||\cdot||_0$ is $l_0$ norm and, physically, is equal to the number of the captured evaders. 
 An optimal controller could be synthesized to optimize the above performance objective, whereas the opponent evasion strategy seeks to reduce the performance objective. Eventually, both sides achieve the well-known Nash equilibrium as a result of the non-cooperative dynamic game.

 As to be shown in the next subsection, the performance objective usually used in differential game is the expected capture time, %which can be related to Eq.~(\ref{e:num}) by 
   \begin{eqnarray}\label{e:expected}
v (s, h) =    \frac{t_f}{\mathfrak{N}}, 
\end{eqnarray}
 \noindent where $s$ denotes the searching strategy, $h$ denotes the hiding strategy, and $v$ is the expected capture time. It is worthwhile to mention that all these symbols are consistent with the pioneering reference \cite{Gal:79SIAM}. Given $h$, a control method should be designed to enable the pursuers follow the optimal $s$.

 \begin{figure}%[!htb]
\center
\begin{subfigure}[]{
\begin{tikzpicture}
	\draw [thick] (-2,0) -- (2,0);
	\draw [thick] (2,0) -- (2,-3.5);
	\draw [thick] (2,-3.5) -- (-2,-3.5);
	\draw [thick] (-2,-3.5) -- (-2,0);
	
%	\draw[green, thick, fill=green] (0,-2) circle (1.1 cm);
	\node [above] at (0,-2.2) {$p_1$};
	\node [above] at (-1.,-1.2) {$e_1$};
	\node [above] at (0.,-1.2) {$e_2$};
	\node [above] at (1.3,-0.7) {$e_3$};

	\draw [dashed, thick] (-0.1,-1.8) -- (-1,-1.1);
	\draw [dashed, thick] (-0.85,-0.95) -- (-0.18,-0.95);
	\draw [dashed, thick] (0.1,-0.95) -- (1.2,-0.6);
\end{tikzpicture}}
\end{subfigure}
\hspace{10mm}
\begin{subfigure}[]{
\begin{tikzpicture}
	\draw [thick] (-2,0) -- (2,0);
	\draw [thick] (2,0) -- (2,-3.5);
	\draw [thick] (2,-3.5) -- (-2,-3.5);
	\draw [thick] (-2,-3.5) -- (-2,0);
	
	\draw[green, thick, fill=green] (0,-2) circle (1.1 cm);
	\draw[green, thick, fill=green, opacity=0.3] (0,-1.1) circle (1.1 cm);
	\node [above] at (0,-2.2) {$p_1$};
	\node [above] at (-1.,-1.2) {$e_1$};
	\node [above] at (0.,-1.2) {$e_2$};
	\node [above] at (1.3,-0.7) {$e_3$};

	\draw [dashed, thick] (-0.1,-1.8) -- (-0.1,-1.15);
	\draw [dashed, thick] (-0.85,-0.95) -- (-0.18,-0.95);
	\draw [dashed, thick] (-1,-0.85) -- (-1.6,-0.5);
	\node [above] at (-1.6,-0.5) {?};
	\draw [dashed, thick] (-1,-1.05) -- (-1.6,-1.5);
	\node [below] at (-1.6,-1.5) {?};
\end{tikzpicture}}
\end{subfigure}
\caption{ The searching strategy (a) with full observations and (b) partial observations (denoted by the green circles).   }
\label{f:control}
\end{figure}
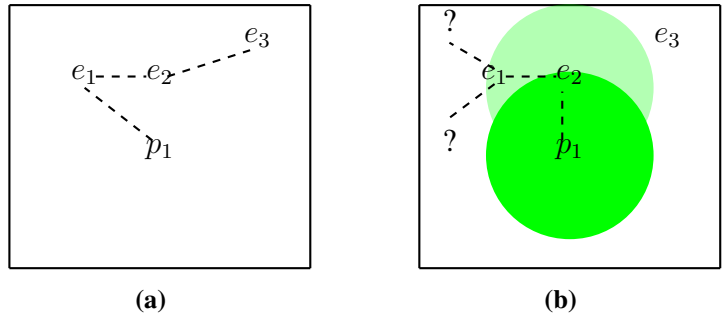
Nevertheless, the StarCraft mini-game set-up can only access to partial observations that prevent an optimal control from possible. To show this, an example simply with 1 pursuer and 3 immobile evaders is conceived in Fig.~\ref{f:control}. When the locations of all three evaders are known to the pursuer, it is easy to see that the optimal searching strategy is as those shown in Fig.~\ref{f:control}(a), first from $p_1$ to $e_1$, and then towards $e_2$ and $e_3$ consecutively. However, when the observation is partial, only $e_2$ is known to $p_1$ in the initial set-up (refer to the green circle). When $p_1$ captures  $e_1$,  the initial invisible $e_2$ becomes to be visible to $p_1$ (refer to the light green circle). Compared to Fig.~\ref{f:control}(a), this searching strategy is certainly not so optimal, not to mention that $e_3$ is still in the fog of war that requests further explorations (represented by the question marks in Fig.~\ref{f:control}(b)).

%The best choice for the two players is a control pair such that no one can benefit by changing his/her control while the other keeps his/hers unchanged... called an open-loop saddle point, whose existence can be proved in  stochastic linear-quadratic differential games \cite{Sun:21siam}, which also considered a much more generic set-up with nonzero state dynamic matrice. 
%A particularly challeng- ing class of problems in this area is partially observable, cooperative, multi-agent learnin, 

\begin{figure}[htbp!]
  \begin{center}
              \subfigure[]{
  \includegraphics[width=5cm]{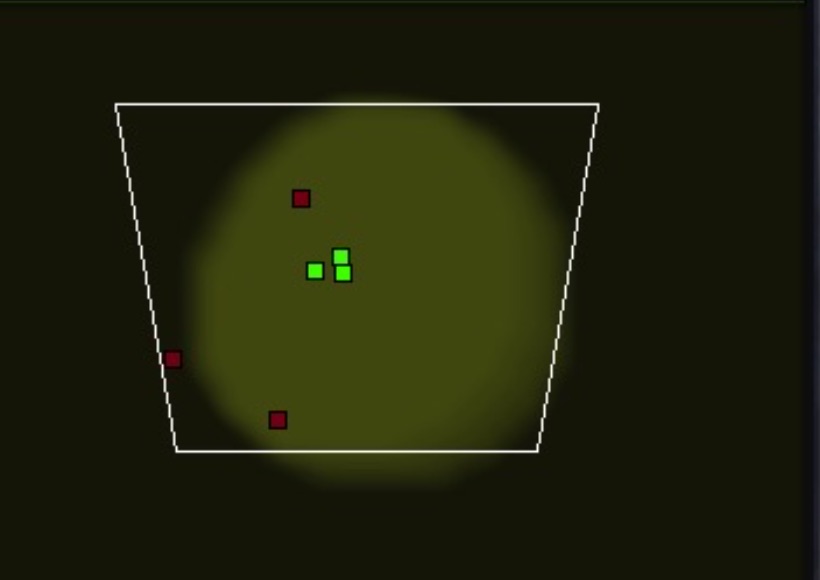}}
	\hspace{10mm}
            \subfigure[]{
  \includegraphics[width=5cm]{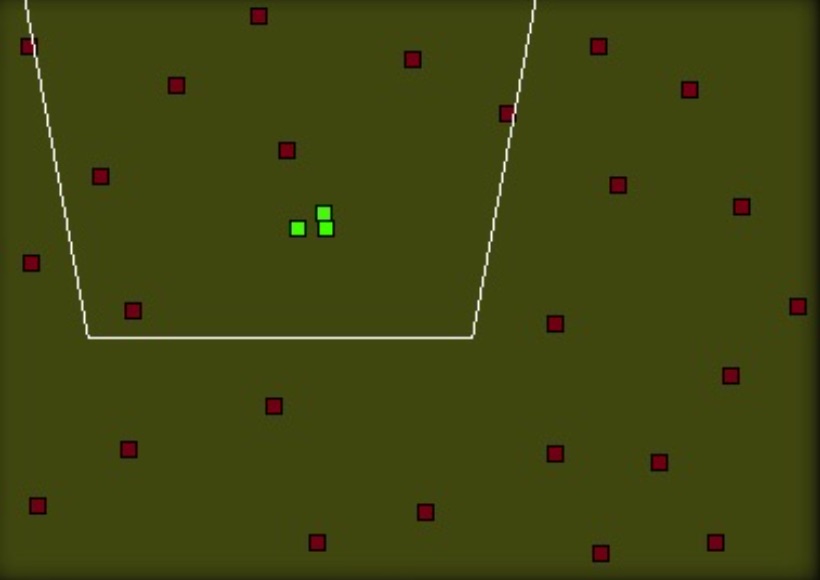}}
  \caption{ The mini-game of FindAndDefeatZerglings when fog of war is (a) activated and (b) deactivated, respectively. }
  \label{f:FBZ}
  \end{center}
\end{figure}
It should be noted that Fig.~\ref{f:control} only considers a much simplified scenario. The FindAndDefeatZerglings mini-game contains more complicated features. Figure~\ref{f:FBZ}(a) shows a classical radar screenshot of the FindAndDefeatZerglings mini-game.   
Figure~\ref{f:FBZ}(b) shows the corresponding screenshot when for of war is deactivated, where the three green dots represent the marine units (i.e., the pursuers) and the 25 red dots represent the zerglings (i.e., the evaders). The mini-game has been designed \footnote{Use the script inside the StarCraft II map editor.} to randomly redeploy 25 new evaders only when the former 25 evaders are all captured, which challenges the exploration capabilities of a searching agent in the presence of fog of war. As a result, the FindAndDefeatZerglings mini-game is the most difficult one in the seven mini-games provided by DeepMind in terms of training cost. The reference \cite{Vinyals:17star} has reported that an AI agent can capture 46 evaders after $600\,$M training steps (refer to Fig.~6 therein).  The other reference \cite{pysc2deep} has reported to  capture 45 after $1260\,$M training steps (i.e., $450\,$K episodes therein). The same test has been performed in this work for a deep Q network (DQN) agent with around $8\,$M trainable parameters on a desktop with decent training hardware (Nvidia GeForce RTX 3090). The training speed is quite slow, at $34\,$M steps per day.  %Overall, the analysis shows that it is difficult, if not impossible, to just use classical control method to handle pursuit--evasion game in the presence of fog of war. Moreover, the reinforcement learning cost will be prohibitive xxxx. 

%\section{Strategy of the pursuer}
\subsection{A differential game perspective}
Theoretically, a differential game consists of two or more players against one another in an adversary environment with competing objectives. The corresponding theoretic studies have produced many classical findings, such as but not limited to the references \cite{Gal:79SIAM,Sun:21siam,Isaacs65}. Such a game theory perspective is adopted here to understand the achievable performance for pursuit--evasion in the current complicated StarCraft mini-game set-up. Readers who are only interested in the reinforcement learning environment can neglect the following theoretical developments and directly jump to the next section. 

%possibly through reinforcement learning. 
%To be consistent with former works,

More specifically, the derivations inside Theorem 3 (e.g., Eqs.~(46), (61) and (62)) from the reference \cite{Gal:79SIAM} are identified to be particularly useful for the current pursuit--evasion game problem. 
By essentially following those derivations (but with different simplifications), a constructive proof of the following theorem can be achieved. 

\begin{figure}%[!htb]
\center
\begin{tikzpicture}
%	\draw[-stealth] (0,0) -- (-3.5,-3.5);
	\draw[-stealth,thick] (-5,0) -- (3.5,0);	
	\draw[-stealth,thick] (-5,0) -- (-5,-5);
	\draw [thick] (-5,0) -- (3,0);
	\draw [thick] (3,0) -- (3,-4.5);
	\draw [thick] (3,-4.5) -- (-5,-4.5);
	\draw [thick] (-5,-4.5) -- (-5,0);

%	\node [left] at (-3.5,-3.5) {$z$};
	\node [left] at (-4.5,-5) {$y$};
	\node [above] at (3.7,0) {$x$};
	\node [left] at (-5,0) {$o$};
	
	\draw [-stealth,dashed, thick, blue] (-4.8,-0.5) -- (2.8,-0.5);
	\draw [-stealth,dashed, thick, blue]  (2.8,-0.5)  -- (2.8,-1.5) ;
	\draw [-stealth,dashed, thick, blue] (2.8,-1.5)  -- (-4.8,-1.5);
	\draw [-stealth,dashed, thick, blue] (-4.8,-1.5)  -- (-4.8,-2.5);
	\draw [-stealth,dashed, thick, blue]  (-4.8,-2.5)  -- (2.8,-2.5) ;
	\draw [-stealth,dashed, thick, blue] (2.8,-2.5)  -- (2.8,-3.5);
	\draw [-stealth,dashed, thick, blue]  (2.8,-3.5) --  (-4.8,-3.5);
	\draw [-stealth,dashed, thick, blue] (-4.8,-3.5)  -- (-4.8,-4.25);
	\draw [-stealth,dashed, thick, blue] (-4.8,-4.25)  -- (2.8,-4.25);
	
	\draw [dashed, thick] (-2.5,0) -- (-2.5,-4.5);
	\draw [dashed, thick] (0.5,0) -- (0.5,-4.5);
	\node [above] at (-3.75,0) {$a$};
	\node [above] at (1.75,0) {$a$};
	\node [below] at (-1,-2) {$\cdots$~$\cdots$};

	\draw [dashed, thick] (-5,-1) -- (-2.5,-1);
	\draw [dashed, thick] (-5,-2) -- (-2.5,-2);
	\draw [dashed, thick] (-5,-3.8) -- (-2.5,-3.8);
	\node [left] at (-5,-1.5) {$c$};
	\node [left] at (-5,-0.5) {$c$};
	\node [left] at (-5,-4.1) {$\bar{c} \leq c$};
	
	\node [below] at (-3.8,-1.9) {$\vdots$};
	\node [below] at (-3.8,-2.6) {$\vdots$};	
	\node [below] at (-5.15,-1.9) {$\vdots$};
	\node [below] at (-5.15,-2.6) {$\vdots$};		

\end{tikzpicture}
\caption{ Sketch of the game problem, where the origin $o$ is set to the top left, and the dotted arrowed curves represnt the traversal path of a possible searching strategy. Here the coordinate system follows the StarCraft environment.   }
\label{f:gamedomain}
\end{figure}
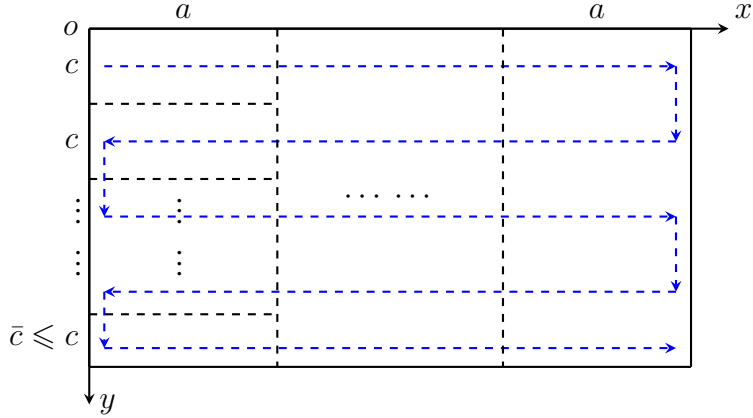

\textbf{Theorem 1.} \textit{There exists an optimal searching strategy $s^*$ in the game domain $Q$ such that for any evading trajectory $h$
used by the evader, the expected capture time $v(s^*, h)$ satisfies}
\begin{equation}
%v(s^*, h) = \frac{\mu}{2r} \frac{(1+\delta)}{U}\left( 1+ \frac{R}{a} + \frac{c}{a} \right). 
v(s^*, h) = l_x \left\lceil \frac{l_y}{2r} \right\rceil \left( 1+ \frac{R}{a} + \frac{2r}{a} \right) \frac{1}{U}. 
\end{equation}
\noindent here $\mu = l_x \times l_y$ is the area of the game domain $Q$, $U$ is the moving speed of the pursuer, 
$R$ is the longest distance between any two points inside the game domain, and $a$ and $c$  
are discretized lengthscales of the 2D game domain (see Fig.~\ref{f:gamedomain}).   
Moreover, $r$ was the sight range of the pursuer in the former work \cite{Gal:79SIAM}, whereas $r$ shall be the attack range in this work and the reason will be given below.   

\textit{Proof.} As shown in Fig.~\ref{f:gamedomain}, $a$ and $c$ are discretizations in the $x$ and $y$ directions. Assume $l_x / a =  \left\lceil {l_x}{/a} \right\rceil  $, where $\left\lceil  \cdot \right\rceil$ is the ceiling function. Then, when  $\bar{c} = c$, the number of small discretized blocks inside the game area $\mu$  is 
\begin{equation}
M=\frac{\mu}{ac}. 
\end{equation}
\noindent Otherwise, when $\bar{c}< c$, 
\begin{equation}\label{e:m}
M  =  \frac{l_x}{a} \left\lceil  \frac{l_y}{c} \right\rceil, %  = \frac{\mu}{ac}(1+\delta), 
\end{equation}
%
%\noindent where $\delta= \left\lceil  \mu/ac \right\rceil - \mu/ac$, with $\left\lceil  \cdot \right\rceil$ the ceiling function. 
\noindent  and $c \leq 2r$ to ensure that pursuers are able to attack any evader when the latter is within the attack range. %, and $\delta= (\left\lceil  \mu/ac \right\rceil - \mu/ac)ac/\mu$, usually much smaller than 1. 

According to \cite{Gal:79SIAM} (especially the proof of Theorem 3 therein),    
during the time segment $0 < t \leq (R + (a+c))/U$, the pursuer could move to any small block of size  $a\times b$, followed by first searching along the horizontal $x$ line and then by searching along the vertical $y$ line.  
On the other hand, when there is only one evader randomly deployed in $M$ blocks, it is easy to see that the probability $p$ of capture satisfies 
\begin{equation}\label{eq:prob}
p = \frac{1}{M}. 
\end{equation}

The pursuer would adopt the search strategy $s^*$ that consists of independent repetitions of the above process for any hiding trajectory $h$. Then, the capture probability ${p}_K$ after the $K\,$th searching with time $t=K(R + (a+c))/U$ satisfies
\begin{equation}\label{e:pk}
{p}_K = (1-p)^K. 
\end{equation}
Hence, from Eqs.~(\ref{e:m})--(\ref{e:pk}) and the Maclaurin series expansion of $1/p$, the expected capture time $v(s^*, h)$ satisfies
\begin{eqnarray}\label{e:theo1}
\nonumber v(s^*, h) &=& \frac{R + (a+c)}{U} \sum_{K=0}^{\infty} {p}_K = \frac{R + (a+c)}{pU} \\ \nonumber
&=&  M\frac{R + (a+c)}{U} = \frac{l_x}{a} \left\lceil \frac{l_y}{c} \right\rceil \frac{R + (a+c)}{U}
\\ &=& {l_x} \left\lceil \frac{l_y}{2r}  \right\rceil \left(1 + \frac{R}{a} + \frac{2r}{a}\right)\frac{1}{U}.    ~\hspace{5mm} \qedsymbol
\end{eqnarray}
%
\iffalse
\begin{eqnarray}\label{e:theo1}
\nonumber v(s^*, h) &=& \frac{R + (a+c)}{U} \sum_{K=0}^{\infty} {p}_K = \frac{R + (a+c)}{pU} \\ \nonumber
&=&  M\frac{R + (a+c)}{U} = \\frac{l_y}{c} \frac{R + (a+c)}{U}(1+\delta) 
\\ &=& \frac{\mu}{2r} \left(1 + \frac{R}{a} + \frac{c}{a}\right)\frac{1}{U}(1+\delta).    ~\hspace{5mm} \qedsymbol
\end{eqnarray}
\fi
%

\textbf{Comment 1.}  The derivation for Theorem 3 of the reference \cite{Gal:79SIAM} was further extensively simplified therein, which is deemed unnecessary for the current game set-up, because each term in Eq.~(\ref{e:theo1}) already holds clear physical meaning.

%To ensure this, $a$ and $c$ mush be equal to $2r$ when evaders are mobile!!! 
%

\textbf{Proposition 1.} \textit{When the evaders are immobile, the
searching strategy can be simplified to a  consecutive traversal. The corresponding expected capture time $v(s^*, h)$ satisfies}
\begin{equation}\label{e:propo1}
v(s^*, h) = l_x \left\lceil \frac{l_y}{2r} \right\rceil \left(1 + \frac{2r}{a}\right)  \frac{1}{U}.   
\end{equation}
\textit{Proof.} For a hiding strategy $h$ with immobile evaders, a consecutive traversal strategy (the dotted arrowed curves in Fig.~\ref{f:gamedomain}) would remove $R$ from Eq.~(\ref{e:theo1}), to directly produce Eq.~(\ref{e:propo1}).  $\qedsymbol$
%
\iffalse
\begin{eqnarray}\label{e:propo1}
 v(s^*, h) & = &  \left\lceil \frac{\mu}{2r}  \right\rceil \left(1 + \frac{c}{a}\right)\frac{1}{U}.   ~\hspace{5mm} \qedsymbol
\end{eqnarray}
%
\fi
%\noindent  

\textbf{Comment 2.} From Eq.~(\ref{e:propo1}), it can be seen a larger $a$ will result in a more rapid capture time. As shown in Fig.~\ref{f:gamedomain}, a possible searching strategy is traversal of the whole game domain, where $a$ could be increased to the whole length in the $x$ direction (that is, $a=l_x=32$). 
Moreover,  $b=2r$ with $r$ the attacking range to ensure that the pursuer would be able to defeat the evader when the latter is visible. 
%As a result, for the current map with size of $32\times 32$, 
%
%\begin{eqnarray}\label{e:delta}
%\delta =  \left\lceil  \frac{32}{2r}\right\rceil  -\frac{32}{2r}. 
%\end{eqnarray}
%

\textbf{Proposition 2.} \textit{The expected capture time for the set-up with $N$ independent evaders is equal to that of the set-up with one evader.} 

\textit{Proof.}  Assume the number of evaders is $N$. From Eq.~(\ref{eq:prob}), the new capture probability will be 
\begin{equation}\label{eq:prob2}
p_N = \frac{N}{M} = N \cdot p 
\end{equation}
\noindent for $N$ evaders. 
Then, from Eq.~(\ref{e:theo1}), the new expected capture time for all $N$ evaders is 
\begin{eqnarray}\label{e:theo2}
 v_N(s^*, h) = N \cdot v(s^*, h)  = N  \frac{R + (a+c)}{p_N U}  =   \frac{R + (a+c)}{p U}  = v(s^*, h) .    ~\hspace{5mm} \qedsymbol
\end{eqnarray}

\textbf{Comment 4.} 
The reward examined in the mini-games is the number of defeated evaders. 
Given the expected capture time $v(s^*, h) $, the reward becomes
\begin{eqnarray}\label{e:cost}
\mathfrak{R} = \frac{T_f - T_k}{v(s^*, h)}\times N_e = \frac{T_f - \frac{N_e H_e}{N_p \mathrm{DPS}} }{v(s^*, h)}\times N_e,
\end{eqnarray}
\noindent where the game finish time is $T_f=180\,$s,  $T_k$ is the time that is required to defeat all the evaders when they are all within attack range,  $N_{e, p}$ is the number of evaders/pursuers, $H_e$ is the health of each evader, and DPS represents damage per second imposed by the selected units. After substituting those unit parameter values (from the appendix) and Eq.~(\ref{e:propo1}) into Eq.~(\ref{e:cost}), the possible best mean score (mean captured number) performance for the FindAndDefeatZerglings mini-game would be
\begin{eqnarray}\label{e:cost2}
\mathfrak{R} = \frac{180 - \frac{25 \times 35}{3 \times 9.8} }{ 32\left\lceil \frac{32}{2 \times 5}  \right\rceil \left(1 + \frac{2 \times 5}{32}\right)\frac{1}{3.15}}\times 25 \approx 70, 
\end{eqnarray}
\noindent which is very close to the best mean score 62 currently achieved by a relational agent from DeepMind \cite{Zambaldi:19relational}. The slight difference could be caused by the effect due to $R$ that has been neglected in the above calculation. The effect could be important especially at the game reset state and its absence in  Eq.~(\ref{e:cost2}) could thus yield a slight overestimation. On the other hand, a further optimization on agent network structures could possibly furhter increase the achievable mean score.  Overall, the game theoretic perspective helps to increase our understanding of the achievable performance for the current StarCraft mini-games.

\section{The StarCraft Adversary-Agent Challenge}\label{s:SAAC}

In the abovementioned FindAndDefeatZerglings mini-game, all evaders are almost immobile. The evaders (here is zergling, see Fig.~\ref{f:units}(b)) will remain still but run towards the pursuers (here is marine, see Fig.~\ref{f:units}(a)) when the latter are visible. Such a hiding strategy actually simplifies the searching task for the pursuers and justifies the simplified assumption adopted in Eq.~(\ref{e:propo1}). Gal has pointed out that a mobile evader is more difficult to be captured  \cite{Gal:79SIAM}, which motivates this work to develop a new adversary-agent learning environment that can be used to train an AI agent for mobile evaders. 

%Hence, a DQN agent controls evaders to manueuver in the proposed new adversary-agent learning environment. Details can be found in the next section.  %the optimal hiding strategy should let is randomly  
% An mobile evader could e

%
 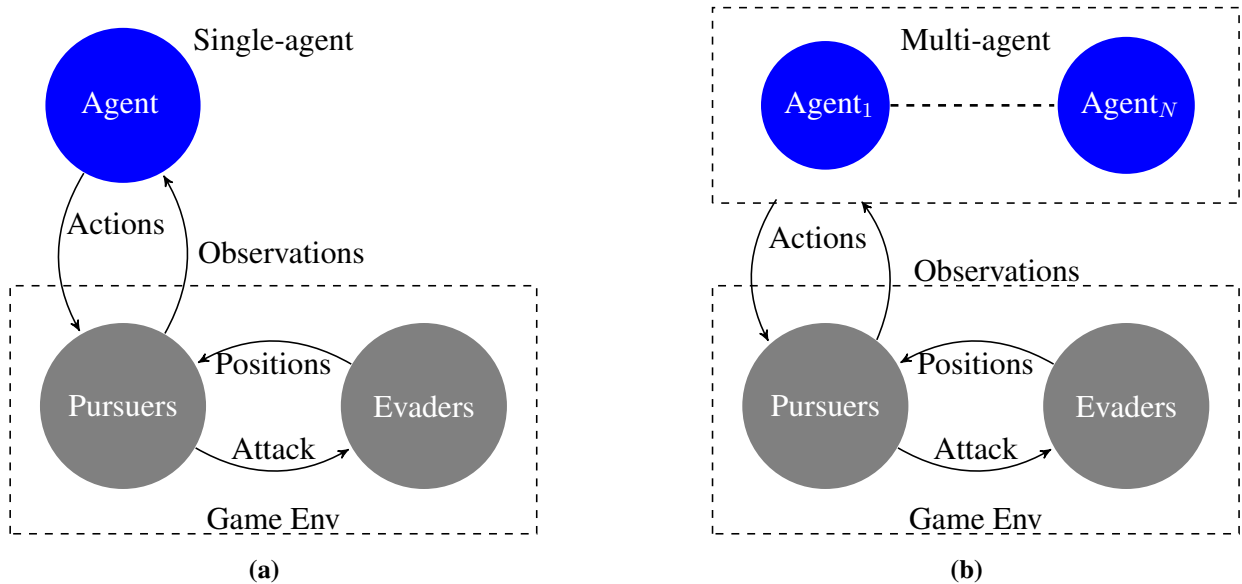
\begin{figure}
             \subfigure[]{
 \begin{tikzpicture}[->,>=stealth',shorten >=1pt,auto,node distance=4cm,
  semithick]
  \tikzstyle{every state}=[fill=gray,draw=none,text=white]

  \node[state] 	    (A)                    {\hspace{1mm}~Pursuers\hspace{2mm}~};
  \node[state,fill=blue]         (B) [above  of=A] {\hspace{1.75mm}~Agent~\hspace{2.6mm}~};
  \node[state]         (C) [right of=A] {\hspace{1.55mm}~Evaders\hspace{2.5mm}~};

  \path (A) 
            edge  [bend right]              node[above=1mm] {\hspace{-20mm} Actions} (B)
            edge [bend right]                node {Attack} (C);
%	(C) edge [loop right] node {Learning} (C)
%	edge  	node{Inversion}(B);
  \path (C) 
            edge [bend right]             node {Positions} (A);

  \path (B) 
            edge  [bend right]               node[] {\hspace{17mm}Observations} (A);
		
	\draw[dashed] (-1.5,-1.7) rectangle (5.5,1.6);
	
	\node[below] (nodea) at (2,-1.2){Game Env};
	\node[above] (nodea) at (2,4.5){Single-agent};
\end{tikzpicture}}
\hspace{20mm}
             \subfigure[]{
 \begin{tikzpicture}[->,>=stealth',shorten >=1pt,auto,node distance=4cm,
  semithick]
  \tikzstyle{every state}=[fill=gray,draw=none,text=white]

  \node[state] 	    (A)                    {\hspace{1mm}~Pursuers\hspace{2mm}~};
  \node[state,fill=blue]         (B) [above  of=A] {~Agent$_1$~};
    \node[state,fill=blue]         (D) [right  of=B] {~Agent$_N$~};
    
    \node[rectangle, dashed] (rec) at (-0.6,2.9) {\hspace{6mm}~~\hspace{6mm}};
    \node[rectangle,  dashed] (rec2) at (0.4,2.9) {\hspace{6mm}~~\hspace{6mm}};
        
  \node[state]         (C) [right of=A] {\hspace{1.55mm}~Evaders\hspace{2.5mm}~};

  \path (A) 
            edge  [bend right]              node[above=1mm] {\hspace{-20mm} Actions} (rec2)
            edge [bend right]                node {Attack} (C);
%	(C) edge [loop right] node {Learning} (C)
%	edge  	node{Inversion}(B);
  \path (C) 
            edge [bend right]             node {Positions} (A);
%            edge  [bend right]              node[above=1mm] {\hspace{-20mm} Actions} (D);
	\draw [line width=1pt,-,black, dashed] (B) -- (D);
  \path (rec) 
            edge  [bend right]               node[] {\hspace{20mm}Observations} (A);
%\path (D) 
%	  edge  [bend right]               node[] {\hspace{17mm}Observations} (C);		
	\draw[dashed] (-1.5,-1.7) rectangle (5.5,1.6);	
	\draw[dashed] (-1.5,2.7) rectangle (5.5,5.3);
	\node[below] (nodea) at (2,-1.2){Game Env};
	\node[above] (nodea) at (2,4.5){Multi-agent};
\end{tikzpicture}}
  \caption{  (a) The learning environment of the mini-map from DeepMind provides an interface to a single agent, which has been extended to (b) multiple cooperative agents in the SMAC learning environment \cite{Samvelyan:19smac}.
  }\label{f:sketch} 
   \end{figure}
First, Fig.~\ref{f:sketch} compares the structures of the existing mini-game reinforcement learning environments.  
The FindAndDefeatZerglings mini-game essentially follows the diagram shown in
Fig.~\ref{f:sketch}(a), where a single agent interacts with StarCraft II environment and  
controls pursuers to maximize future rewards, whereas the build-in script code from StarCraft II controls evaders (to either remain still or push back). The SMAC toolkit \cite{Samvelyan:19smac} follows the diagram shown in Fig.~\ref{f:sketch}(b), where actions from each agent are concatenated through the SMAC toolkit, and observations from StarCraft environment are separated and redistributed to each of the multi-agents. 
The paradigm underneath is  a centralized training but decentralized execution \cite{Kraemer:16}.   
Hence, the SMAC toolkit enables reinforcement learning of coordinated actions within multiple cooperative agents. However, as far as this author knows, an adversary-agent environment that would enable reinforcement learning, especially for pursuit--evasion type differential game, is still rare. To fill this gap, 
the current work endeavors to propose an adversary-agent learning environment (was named StarCraft Adversary-Agent Challenge, SAAC). % by extending the StarCraft mini-game set-ups.   

 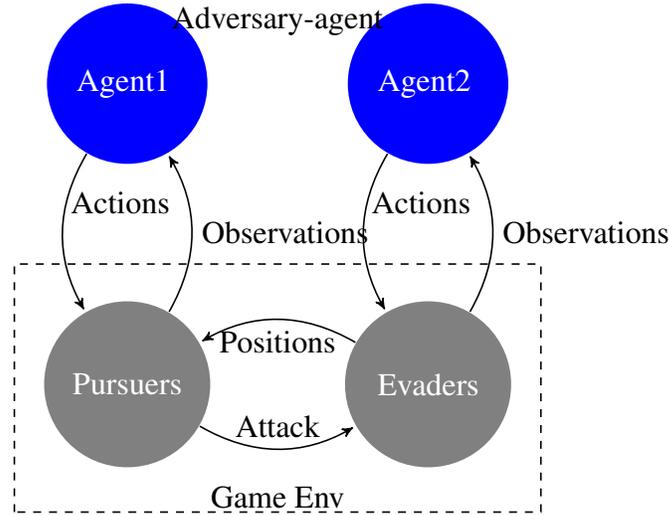
\begin{figure}
 \center
 \begin{tikzpicture}[->,>=stealth',shorten >=1pt,auto,node distance=4cm,
  semithick]
  \tikzstyle{every state}=[fill=gray,draw=none,text=white]

  \node[state] 	    (A)                    {\hspace{1mm}~Pursuers\hspace{2mm}~};
  \node[state,fill=blue]         (B) [above  of=A] {\hspace{1.mm}~Agent1~\hspace{2.mm}~};
    \node[state,fill=blue]         (D) [above  of=C] {\hspace{1.mm}~Agent2~\hspace{2.mm}~};
  \node[state]         (C) [right of=A] {\hspace{1.55mm}~Evaders\hspace{2.5mm}~};

  \path (A) 
            edge  [bend right]              node[above=1mm] {\hspace{-20mm} Actions} (B)
            edge [bend right]                node {Attack} (C);
%	(C) edge [loop right] node {Learning} (C)
%	edge  	node{Inversion}(B);
  \path (C) 
            edge [bend right]             node {Positions} (A)
            edge  [bend right]              node[above=1mm] {\hspace{-20mm} Actions} (D);

  \path (B) 
            edge  [bend right]               node[] {\hspace{17mm}Observations} (A);

\path (D) 
	  edge  [bend right]               node[] {\hspace{17mm}Observations} (C);	
	\draw[dashed] (-1.5,-1.7) rectangle (5.5,1.6);
	\node[below] (nodea) at (2,-1.2){Game Env};
		\node[above] (nodea) at (2,4.5){Adversary-agent};
\end{tikzpicture}
  \caption{  Overview of the SAAC learning environment.
  }\label{f:sketch-aa} 
   \end{figure}
Figure~\ref{f:sketch-aa} shows the corresponding structure of the proposed SAAC environment, where two adversary agents control pursuers and evaders, respectively. It is worthwhile to mention that  both agents could be further extended to concatenate multiple coordinating agents by further incorporating SMAC toolkit.

The SAAC environment consists of some example mini-maps and adversary agents, which will guide interested readers to build up their own maps and agents. Some of the findings that are important for the correct implementation of the environment are summarized as follows. 
\begin{itemize}
\item For unknown reasons, the seven mini-game maps from DeepMind cannot support two adversary-agents for opponent players. Hence, in this work, the mini-map is built from scratch by StarCraft map editor. Then, interested readers can download and further edit my mini-map for their own target research problems.    
\item It is well known in the StarCraft programming community that the current PySC2 interface could produce websocket errors during the low-level message passing between multiple agent interfaces. To bypass this issue, a thorough programming debug has been conducted in this work to identify the corresponding code. Then, a temporary fix has been adopted to rectify the issue before any official fix is available from DeepMind in the near future. 
\end{itemize}

\begin{figure}[htbp!]
  \begin{center}
  \includegraphics[height=5cm]{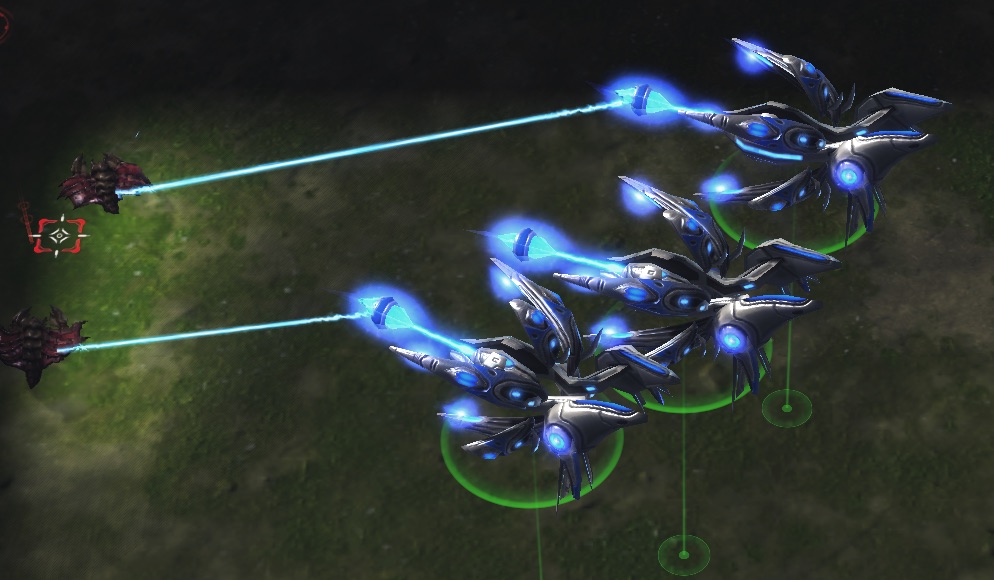} %} 
  \caption{ The screenshot of the adversary-agent learning environment (the FindAndDefeatDrones mini-game) developed in this work. }
  \label{f:newmap}
  \end{center}
\end{figure}
Other important modifications include the modified optimization objectives (to optimize the number of defeated units for pursuers and the number of living units for evaders) and the use of different unit types to address the third issue that has been mentioned in Sec.~\ref{s:SC2}. More specifically, the evaders are changed from Zerg zergling to Zerg drones, which 
are farming workers and will only escape to the nearby fog of war rather than pushing back when they are attacked. The pursuers are changed from Terran marines to  Protoss void ray, which represents a classical type of attack aircraft. Figure~\ref{f:newmap} shows the screenshot of this new, so-called FindAndDefeatDrones mini-game. Compared to the former FindAndDefeatZerglings mini-game, it can be seen that the FindAndDefeatDrones mini-game is more similar to the classical pursuit--evasion game. 
Moreover, other units can be considered in later studies. For example, the set-up with Terran 
medivac dropship versus Protoss void ray shall be able to imitate aerospace interception and capture applications. The corresponding game modifications should be straightforward based on the proposed FindAndDefeatDrones mini-game.

An analysis similar to Eq.~(\ref{e:cost2}) can be conducted for this new mini-game, which yields
%For the proposed new game, if we directly use xxx  the best reward would become
%
\begin{eqnarray}\label{e:cost2}
\nonumber \mathfrak{R} = \frac{180 - \frac{25 \times 40}{3 \times 16.8} }{ 32\left\lceil \frac{32}{2 \times 6}  \right\rceil \left(1 + \frac{2 \times 6}{32}\right)\frac{1}{3.85}}\times 25 \approx 118. 
\end{eqnarray}
\noindent  However, after running this mini-map, an expert human player suggested that the above value is extensively overestimated. It is because that  Eq.~(\ref{e:cost2}) is only 
for immobile evaders (recall the evaders will actually run towards the pursuers in the former mini-game). However, in this new FindAndDefeatDrones mini-game, the evaders (Zerg drones) will escape to the nearby fog of war to avoid to be attacked. Moreover, the moving speed of the Zerg drones is slightly faster than the moving speed of the pursuers (Protoss void ray). 
 The effect of $R$, which is longest possible distance inside the game domain and equal to $\sqrt{2} l_x$, cannot be neglected anymore.  Hence, Eq.~(\ref{e:theo1}) is adopted to yield a new estimation of the expected number of captured units,  
\begin{eqnarray}\label{e:cost2}
\nonumber \mathfrak{R} = \frac{180 - \frac{25 \times 40}{3 \times 16.8} }{ 32\left\lceil \frac{32}{2 \times 6}  \right\rceil \left(1 + \frac{1.4\times32}{32}+ \frac{2 \times 6}{32}\right)\frac{1}{3.85}}\times 25 \approx 60. 
\end{eqnarray}

Before jumping to the learning of the adversary-agent, a couple of tests with simplified scripted agent and random agent \footnote{Interested readers can download SAAC code and type terminal command: python TestScripted\_V2.py} have been conducted to verify and validate the code and the new mini-game set-ups. When the evader agent is random, the testing pursuit agent can achieve mean score of 50.4 (i.e., the number of the captured units), which shows that the whole adversary-agents learning environment is working. Moreover, the pursuit agent is also tested for the classical FindAndDefeatZerglings  mini-game  \footnote{Interested readers can download SAAC code and type terminal command: python TestScripted\_V1.py}  and achieved mean score of 40. Both tests clearly suggest the effectiveness of this testing pursuit agent.

\begin{figure}[htbp!]
  \begin{center}
 %           \subfigure[]{
 % \includegraphics[width=15cm]{figures/Struct1.jpg}}
 %           \subfigure[]{
  \includegraphics[width=11cm]{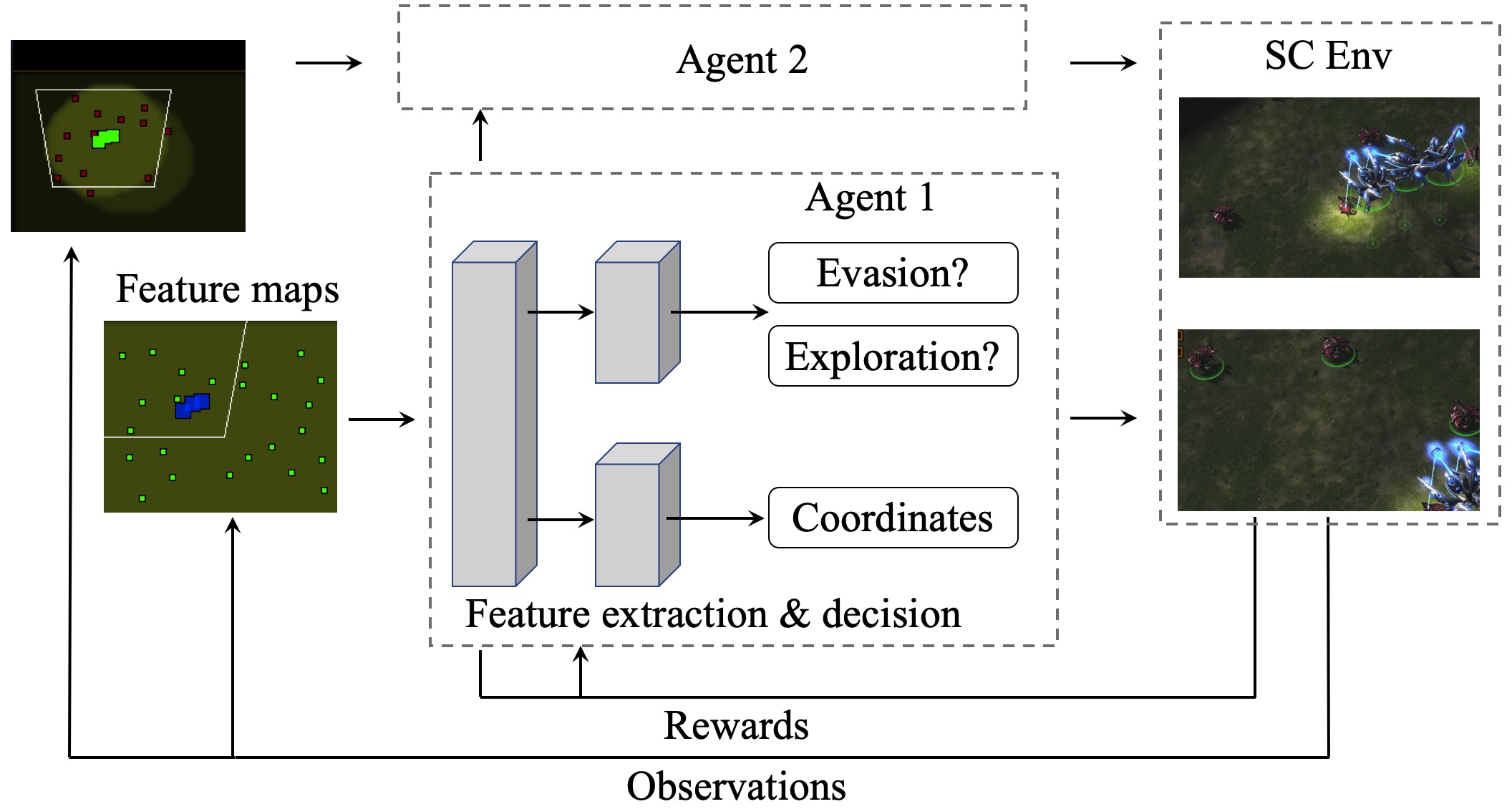} %}
  \caption{ The code structure of the adversary-agents tests for the FindAndDefeatDrones mini-game, where
  a full connected network is adopted in agent 1 to extract features and produce decisions.  }
  \label{f:strut2}
  \end{center}
\end{figure}
Next, the proposed adversary-agents environment is utilized to train agents. To the best knowledge of this author, most former works are focused on pursuit agents  based on A2C,  A3C,  DQN and relational-based neural network methods, but the other side of the coin is rarely studied. Enabled by the new learning environment, here the attention is focus on the training of a pursuit agent. Figure~\ref{f:strut2} shows the code structure, where two interfaces from the StarCraft environment 
output observations (feature maps, etc.) to pursuers and evaders, respectively. Currently, the evader agent (agent 1 in the figure) adopts a four-layers, fully-connected convolutional network architecture. Other hyperparameter values can be found inside the code. The current work only uses such a network to rapidly showcase the proposed adversary learning environment. Further optimizations of the network architecture and hyperparameter configurations can be straightforwardly performed by interested readers.  The pursuit agent (agent 2 in Fig.~\ref{f:strut2}) adopts the above-mentioned traversal agent.  Again, this agent can be easily replaced with other reinforcement learning agents.

%, etc.. To be different from those former works, an agent is trained here for the evader player during the reinforcement learning of the adversary environment. 

%The pursuit agent simply adopts an existing one with decent performance in the FindAndDefeatDrones mini-game when the evader agent is disabled, which will behave like the traditional FindAndDefeatZerglings mini-game, and the performance is 50.4 (mean score) and 54 (maximum). 

\begin{figure}[htbp!]
  \begin{center}
%            \subfigure[]{
 % \includegraphics[height=5cm]{figures/MoveToBeacon1.jpg}}
              \subfigure[]{
  \includegraphics[height=11cm]{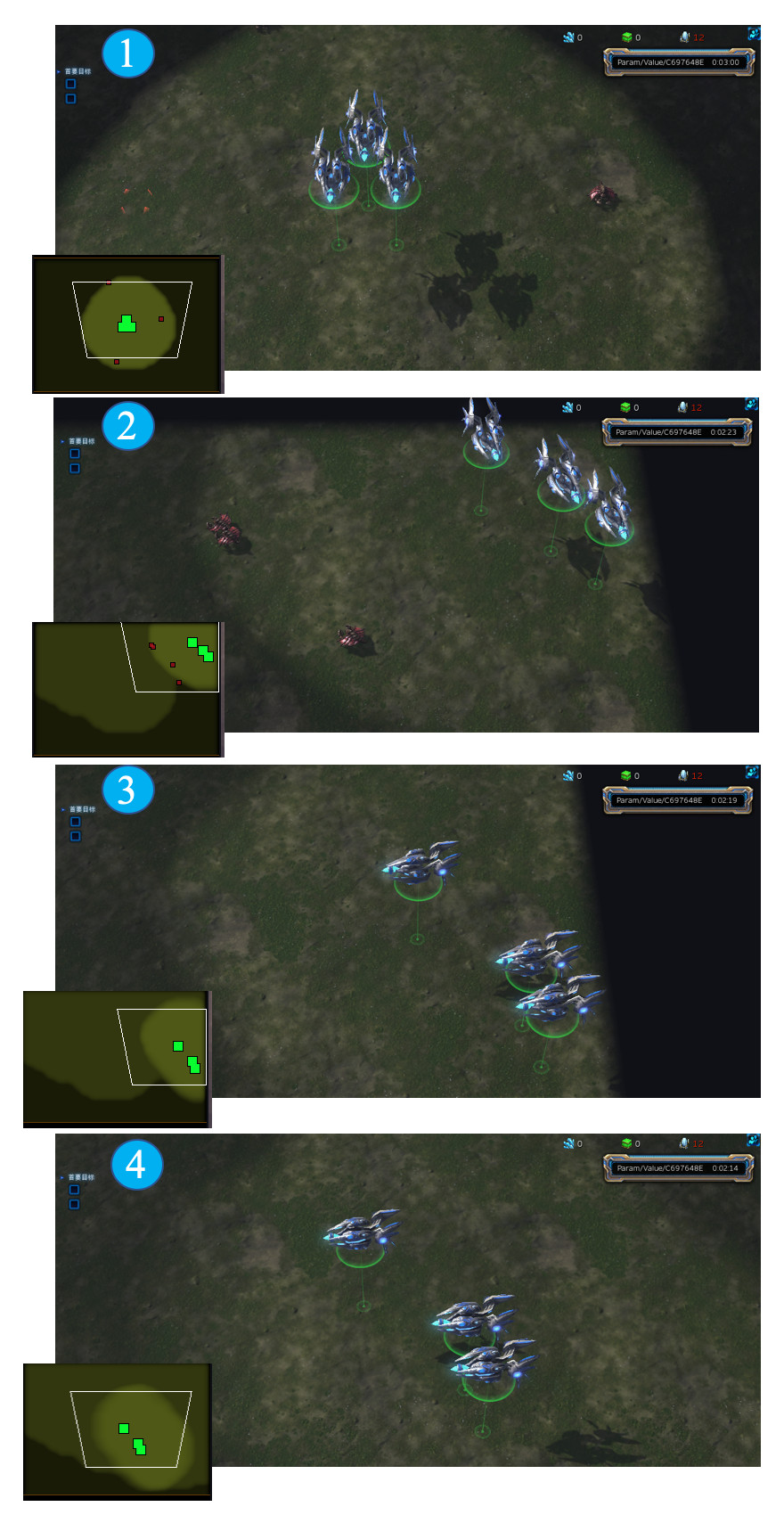}}
  \hspace{5mm}
               \subfigure[]{
  \includegraphics[height=11cm]{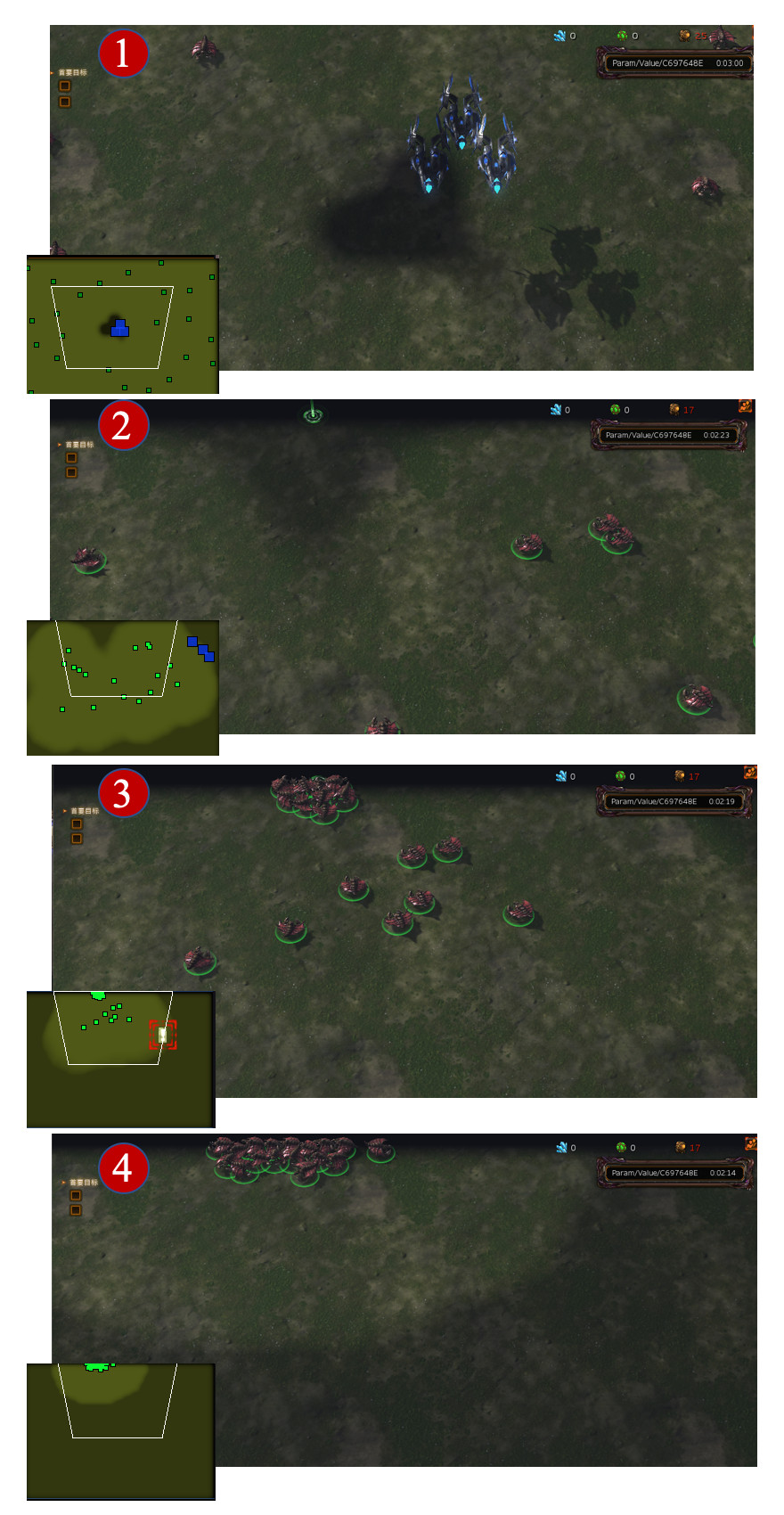}} 
  \\
   \subfigure[]{
    \includegraphics[width=7cm]{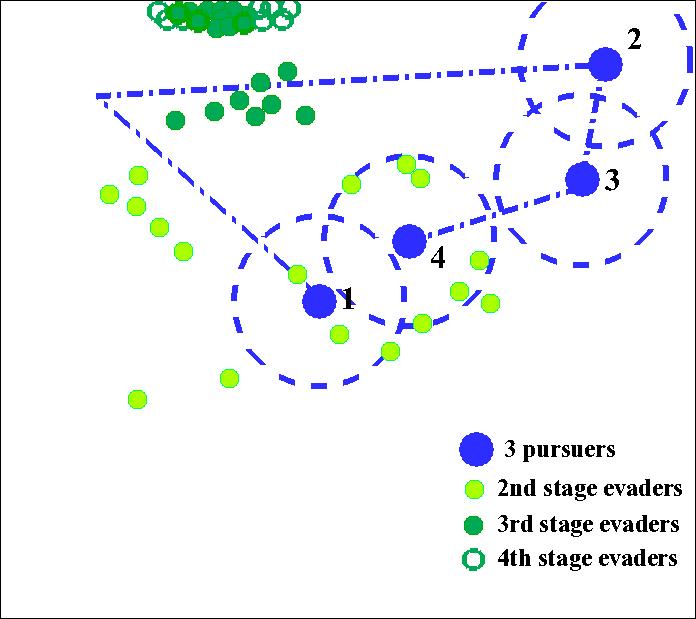}} 
  \caption{  Some of the representative screenshots from (a) the pursuers and (b) the evaders, respectively, during one episode of the FindAndDefeatDrones game. (c) The corresponding searching and evasion strategies, where the dashed circles represent the corresponding attack radius and the dashed lines represent searching paths.      }
  \label{f:FDdrone}
  \end{center}
\end{figure}

Figure~\ref{f:FDdrone} shows some representative screenshots from (a) the pursuers and (b) the evaders, respectively, during one episode of the FindAndDefeatDrones game \footnote{Interested readers can download SAAC code and type terminal command: python exec\_2agents.py.}.
The four consecutive stages are from the initial step to the middle period of a traversal type searching. 
Figure~\ref{f:FDdrone}(c) shows the trajectories of the pursuers in these four stages, and further shows the corresponding spatial distributions (from stage 2 to stage 4) of evading survivals. 
At the first stage, the initial 25 evaders are randomly scattered throughout the whole game domain and, for clarity, are not shown in Fig.~\ref{f:FDdrone}(c). It can be seen that all three pursuers stayed together 
for concentrated firing capability during the searching of the evaders. Similarly, just after a dozen of training episodes, the evader agent learns to control all 25 evaders to gradually move together and eventually convene at either corner of the game domain. 

\begin{figure}[htbp!]
  \begin{center}
  \includegraphics[width=8cm]{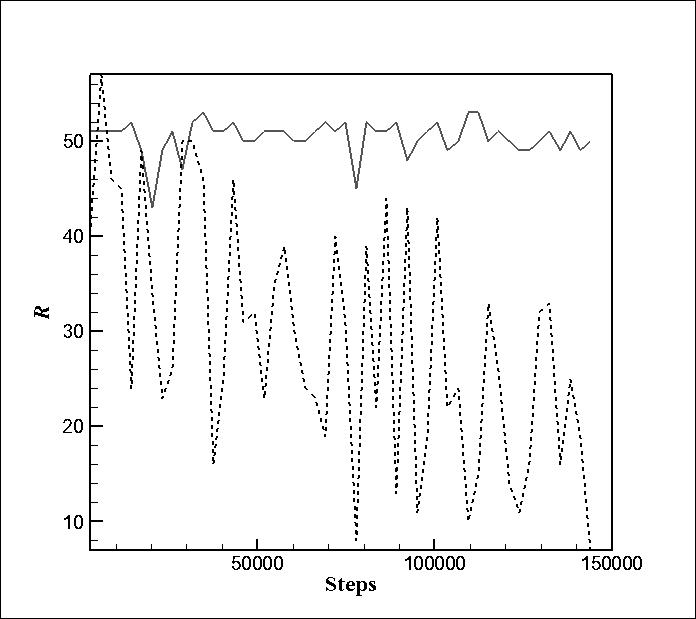}
  \caption{  The number of the captured units for the FindAndDefeatDrones game, where (--) denotes the results achieved by the random agent for the evader part, and ($--$) shows the results through the adversary-agent reinforcement learning with a convolutional network and DQN method.   }
  \label{f:reward}
  \end{center}
\end{figure}

As shown in Fig.~\ref{f:strut2}, the current evader agent only supports collective evasion or collection exploration, which extensively simplified the size of action space and reduce the reinforcement learning time. 
Such a team action strategy can be modified by changing the available action space. 
The game reward $\mathfrak{R}$ shows in 
Fig.~\ref{f:reward} suggests that the reinforcement learning quickly helps to reduce the number of the captured evaders from around 51 (the solid line in the figure) to around 30 (the dashed line). 

Theoretically, through the current group searching and team hiding strategies, the FindAndDefeatDrones mini-game with 3 pursuers and 25 evaders is actually reduced to the classical one princess and one monster game. From Fig.~\ref{f:FDdrone}(c), the hiding strategy learning in the current adversary game environment is similar to the well-known solution from Gal \cite{Gal:79SIAM}, that is, all evaders behavior as one unit,  keep 
moving to a random location as a team and stay still for a certain time interval, and then repeat such a procedure. 
It is worthwhile to mention that the two strategies also imitate the possible action behavior from ordinary human players, who are tended to control a group of units together.  Whether a separate searching or a separate evasion would lead to better rewards is still an interesting open question that request further study, which however is beyond the scope of the current paper.

\section{Conclusion} \label{s:cons}

In this paper, a StarCraft based reinforcement learning environment that supports adversary agents
has been proposed for the study of pursuit--evasion game in the presence of fog of war. 
The key contribution includes the analysis of the potential performance of an agent in the current pursuit--evasion mini-game, by merging control and differential game theory for the specific reinforcement learning problem set-ups, and the development of SAAC environment by extending the current StarCraft mini-games. 
The current work is solely focused on the evader agent learning, which is rare in the former studies, and configures the pursuit agent to a testing traversal agent with decent searching performance. The proposed SAAC environment should also be applicable to the future studies that wish to train adversary agents simultaneously, and the bottleneck that the author can currently envision is the prohibitive training cost.

Theoretically, the most critical part of this work is the analytical explanation of the potential pursuit agent performance by differential game theory for the StarCraft mini-game set-ups. In addition, the resultant performance values help to examine the performance of the traversal pursuit agent used in the adversary-agent trainings. The subsequent study showcases the use of this learning environment and the effectiveness of the learned adversary agent for evasion units. On the other hand, reinforcement learning usually assumes a stationary environment, which could be inapplicable to pursuit--evasion when non-cooperative game dynamics appear.  Hence, the proposed SAAC environment should enable new research directions for both differential game research community  and reinforcement learning research community, and help to promote the merging of both game theory and AI technology together. 

%, where game theory models interactive decision making and control systems methods address evolutionary dynamics while mitigating the uncertainty inherent in human decision making.  

 Last but not least, the author wishes to emphasize that this paper serves as an introduction with a focus especially on the development of the SAAC environment, with detailed explanations of why to design in such a way and how to bypass the inherent code issues and certain software limitations, etc.. The corresponding SAAC code can be found at  GitHub:  https://github.com/xunger99/SAAC-StarCraft-Adversary-Agent-Challenge. More studies regarding different AI network architectures and hyperparameter optimization will be given in the follow-up articles.

%xxx utilize the rapidly developing technologies of deep learning and  help to provide new technological directions for pursuit-and-evasion applications, especially in the presence of realistic limitations, such as rough terrain effects, limited sight and attacking ranges. 

%Some important observations gained from xxx are summarized here.   
% the corresponding tutorial code can be found at GitHub.  
 
 %The corresponding findings should benefit researchers to utilize the StarCraft II environment for their own study of pursuit--evasion problems. ...paving the way for the adoption of GNNs in other fields of science...

%

\iffalse
1. Fog  of war actually helps pursuer. Moreover, I wish to mention that if the eavders are allowed to push back, its sight range almost covers the whole map, and equal that fog of war is disabled. Then, if there is an agent, rather than the build-in bot code, the 25 evaders can stay together and easily defeated the 3 pursuers. 

2. pursuit--evasion whit Fog  of war  = Traversal

3. immobile or escaping? 

4. Compared to those PE problems usually considered in the literature, the games of StarCraft II 
have a much bigger state and action spaces, real-time component, partial observability, and longer game duration, this is a much more difficult task than any attempted before. 
\fi

%CNNs have the ability to extract multi-scale localized spatial features and compose them to construct highly expressive representa- tions, which led to breakthroughs in almost all machine learning areas and started the new era of deep learning (LeCun et al., 2015)....
%

\section*{Acknowledgement}
%
%This research was supported by the NSF Grant of China (Grant No. 11172007), AVIC Commercial Aircraft Engine Company, and SRF for ROCS, SEM.
This research was conducted during the pandemic era when financial resource and student support were both scarce. The author does wish to acknowledge the great affection, emotional support and understanding from his family. 
%

%\bibliographystyle{aiaa}
%\bibliography{references}%

\begin{thebibliography}{10}
\newcommand{\enquote}[1]{``#1''}

\bibitem{Pachter:19AIS}
Pachter, M., Von~Moll, A., Garcia, E., Casbeer, D., and Milutinović, D.,
  \enquote{Cooperative Pursuit by Multiple Pursuers of a Single Evader,} {\em
  Journal of Aerospace Information Systems\/}, Vol.~17, No.~8, 2019,
  pp.~371--389.

\bibitem{Zadka:20jgcd}
Zadka, B., Tripathy, T., Tsalik, R., and Shima, T., \enquote{Consensus-Based
  Cooperative Geometrical Rules for Simultaneous Target Interception,} {\em
  Journal of Guidance, Control, and Dynamics\/}, Vol.~43, No.~12, 2020,
  pp.~2425--2432.

\bibitem{Shen:18jgcd}
Shen, H.~X. and Casalino, L., \enquote{Revisit of the Three-Dimensional Orbital
  Pursuit-Evasion Game,} {\em Journal of Guidance, Control, and Dynamics\/},
  Vol.~41, No.~8, 2018, pp.~1820--1820.

\bibitem{Gutman:19jgcd}
Gutman, S., \enquote{Exoatmospheric Interception via Linear Quadratic
  Optimization,} {\em Journal of Guidance, Control, and Dynamics\/}, Vol.~42,
  No.~3, 2019, pp.~624--631.

\bibitem{Ye:20AST}
Ye, D., Shi, M.~M., and Sun, Z.~W., \enquote{Satellite Proximate
  Pursuit--Evasion Game with Different Thrust Configurations,} {\em Aerospace
  Science and Technology\/}, Vol.~99, No.~4, 2020, pp.~105715(1--10).

\bibitem{Venigalla:21jgcd}
Venigalla, C. and Scheeres, J.~D., \enquote{Delta-$V$-Based Analysis of
  Spacecraft Pursuit--Evasion Games,} {\em Journal of Guidance, Control, and
  Dynamics\/}, 2021, In press.

\bibitem{Leone:21}
Leone, P., Buwaya, J., and Alpern, S., \enquote{Search-and-Rescue Rendezvous,}
  {\em European Journal of Operational Research\/}, 2021, In press.

\bibitem{Gal:79SIAM}
Gal, S., \enquote{Search Games with Mobile and Immobile Hider,} {\em SIAM Journal of Control and Optimization\/}, Vol.~17, No.~1, 1979, pp.~99--122.

\bibitem{Sun:21siam}
Sun, J.~R., \enquote{Two-person Zero-sum Stochastic Linear-quadratic
  Differential Games,} {\em SIAM Journal of Control and Optimization\/},
  Vol.~59, No.~3, 2021, pp.~1804--1829.

\bibitem{Car:18jgcd}
Carr, R.~W., Cobb, R.~G., Pachter, M., and Pierce, S., \enquote{Solution of a
  Pursuit--Evasion Game Using a Near-Optimal Strategy,} {\em Journal of
  Guidance, Control, and Dynamics\/}, Vol.~41, No.~4, 2018, pp.~841--850.

\bibitem{Cao:20nsr}
Cao, M., \enquote{Merging Game Theory and Control Theory in the Era of AI and
  Autonomy,} {\em National Science Review\/}, Vol.~7, No.~7, 2020,
  pp.~1122--1124.

\bibitem{Wang:20neur}
Wang, Y.~D., Dong, L., and Sun, C.~Y., \enquote{Cooperative Control for
  Multi-Player Pursuit--Evasion Games with Reinforcement Learning,} {\em
  Neurocomputing\/}, Vol.~412, 2020, pp.~101--114.

\bibitem{Wang:20ACA}
Wang, X.~Q., Xuan, S.~Z., and Ke, L.~J., \enquote{Cooperatively Pursuing a
  Target Unmanned Aerial Vehicle by Multiple Unmanned Aerial Vehicles based on
  Multiagent Reinforcement Learning,} {\em Advanced Control for
  Applications\/}, Vol.~2, 2020, pp.~e27(1--13).

\bibitem{Li:18PE}
Li, Z., Meyer, N.~J., Laber, E.~B., and Brigantic, R., \enquote{Thompson
  Sampling for Pursuit--Evasion Problems,} arXiv:1811.04471v1, 2018.

\bibitem{PySC2}
DeepMind, \enquote{PySC2 - StarCraft II Learning Environment,}
  https://github.com/deepmind/pysc2.

\bibitem{Samvelyan:19smac}
Samvelyan, M., Rashid, T., and de~Witt, C.~S., et~al., \enquote{The StarCraft
  Multi-Agent Challenge,} arXiv:1902.04043v5, 2019.

\bibitem{Arulkumaran:AlphaStar}
Arulkumaran, K., Cully, A., and Togelius, J., \enquote{AlphaStar: An
  Evolutionary Computation Perspective,} arXiv: 1902.01724v3, 2019.

\bibitem{Vinyals:19nature}
Vinyals, O., Babuschkin, I., and Czarnecki, W.~M., et~al., \enquote{Grandmaster
  Level in StarCraft II using Multi-agent Reinforcement Learning,} {\em
  Nature\/}, Vol.~575, 2019, pp.~350--354.

\bibitem{alphastar_cost}
Wang, K., \enquote{DeepMind achieved StarCraft II GrandMaster Level, but at
  what cost?}
  https://medium.com/swlh/deepmind-achieved-starcraft-ii-grandmaster-level-but-at-what-cost-32891dd990e4.

\bibitem{Vinyals:17star}
Vinyals, O., Ewalds, T., and Bartunov, S., et~al., \enquote{StarCraft II: A New
  Challenge for Reinforcement Learning,} arXiv: 1708.04782v1, 2017.

\bibitem{Zambaldi:19relational}
Zambaldi, V., Raposo, D., and Santoro, A., et~al., \enquote{Deep Reinforcement
  Learning with Relational Inductive Biases,} ICLR, 2019.

\bibitem{Alghanem:A3C18}
Alghanem, B. and Keerthana, P.~G., \enquote{Asynchronous Advantage Actor-Critic
  Agent for Starcraft II,} arXiv:1807.08217v1, 2018.

\bibitem{reaver}
\enquote{Reaver: Modular Deep Reinforcement Learning Framework,}
  https://github.com/inoryy/reaver.

\bibitem{pysc2deep}
\enquote{PySC2 Deep RL Agents,}
  https://github.com/simonmeister/pysc2-rl-agents.

\bibitem{Isaacs65}
Isaacs, R., {\em Differential Games\/}, John Wiley and Sons, 1965.

\bibitem{Bernhard09}
Bernhard, P., Gaitsgory, V., and Pourtallier, O., {\em Annals of the
  International Society of Dynamic Games: Analytical and Numerical
  Developments\/}, Birkhäuser Boston, 2009.

\bibitem{gym}
DeepMind, \enquote{Gym,} https://gym.openai.com.

\bibitem{Hasselt:15DQN}
van Hasselt, H., Guez, A., and Silver, D., \enquote{Deep Reinforcement Learning
  with Double Q-learning,} arXiv:1509.06461v3, 2015.

\bibitem{Shneydor98}
Shneydor, N.~A., {\em Missile Guidance and Pursuit: Kinematics,Dynamics and
  Control\/}, Woodhead Publ., Cambridge, England, U.K., 1998.

\bibitem{Kraemer:16}
Kraemer, L. and Banerjee, B., \enquote{Multi-agent Reinforcement Learning as a
  Rehearsal for Decentralized Planning,} {\em Neurocomputing\/}, Vol.~190,
  2016, pp.~82--94.

\bibitem{SCunits}
\enquote{Legacy of the Void,}
  https://liquipedia.net/starcraft2/Legacy\_of\_the\_Void.

\end{thebibliography}

\appendix
\section*{Appendix}
\subsection{The units}
\renewcommand{\thefigure}{A\arabic{figure}}
\renewcommand{\thetable}{A\arabic{table}}
\setcounter{figure}{0}
\setcounter{table}{0}

\begin{figure}[htbp!]
  \begin{center}
            \subfigure[]{
  \includegraphics[width=4cm]{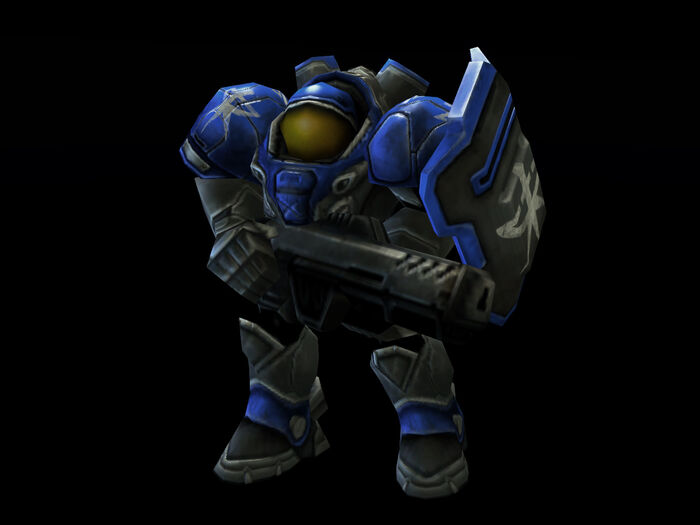}}
            \subfigure[]{
  \includegraphics[width=4cm]{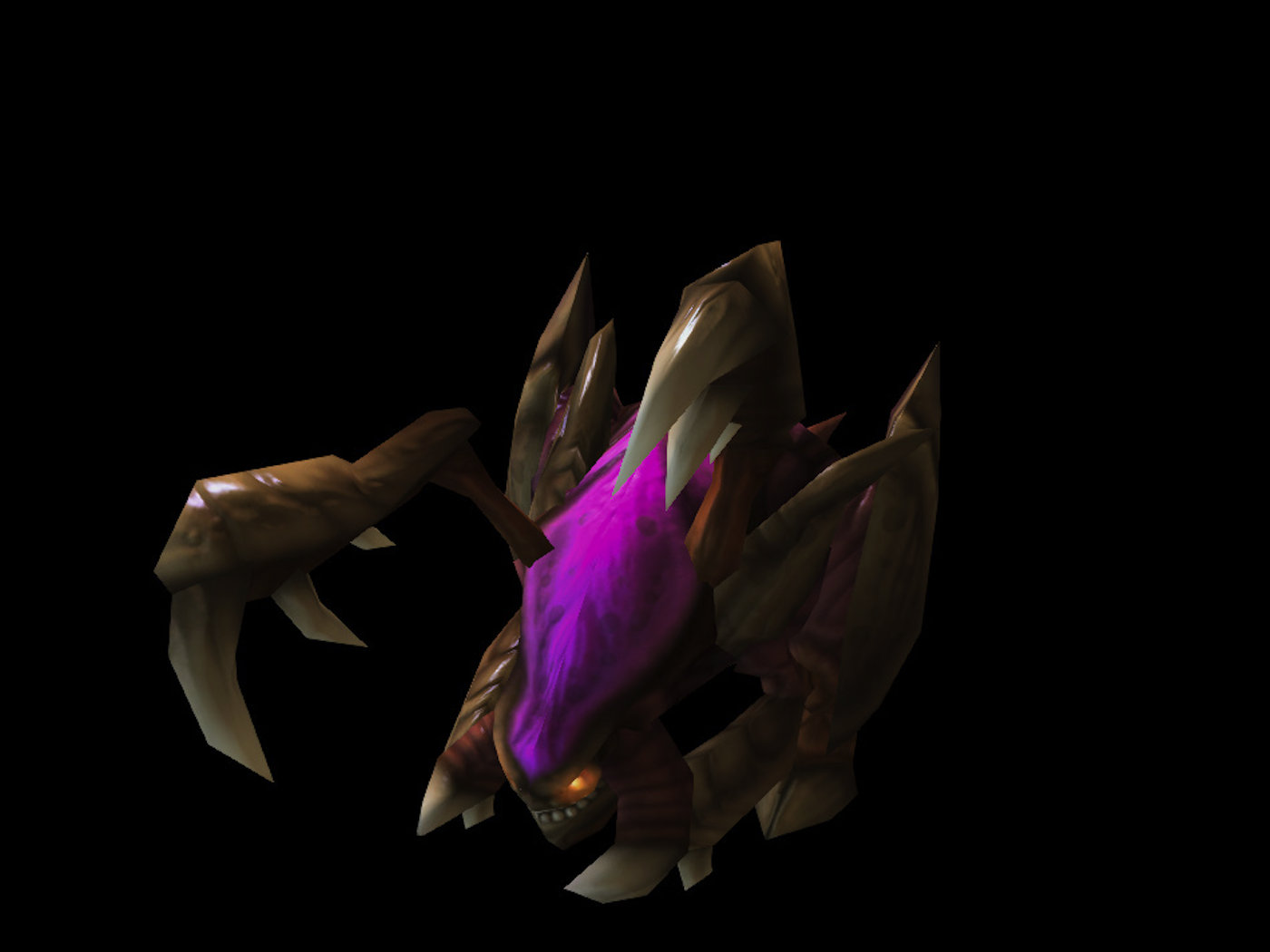}}
             \subfigure[]{
  \includegraphics[width=4cm]{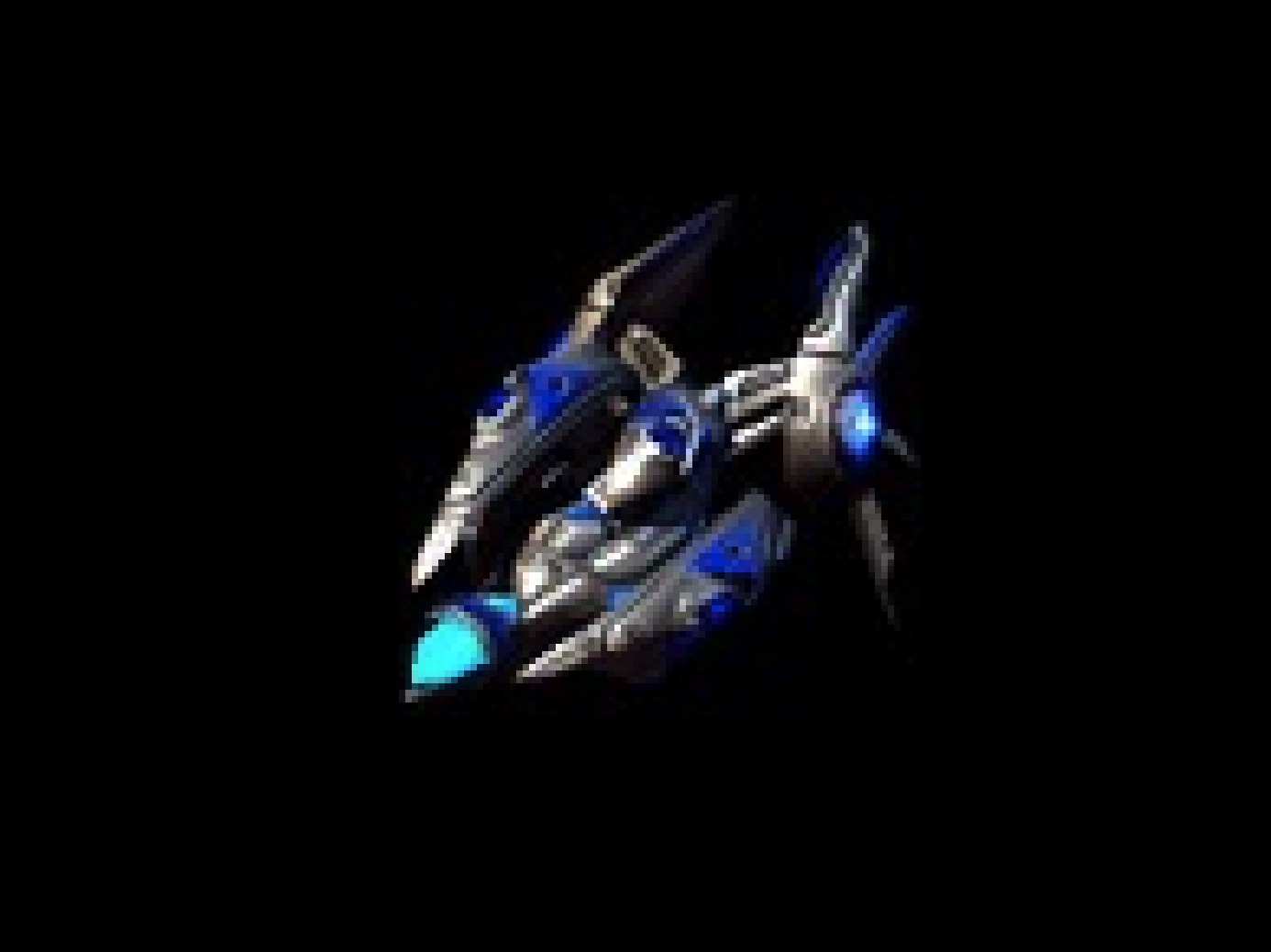}} 
              \subfigure[]{
  \includegraphics[width=4cm]{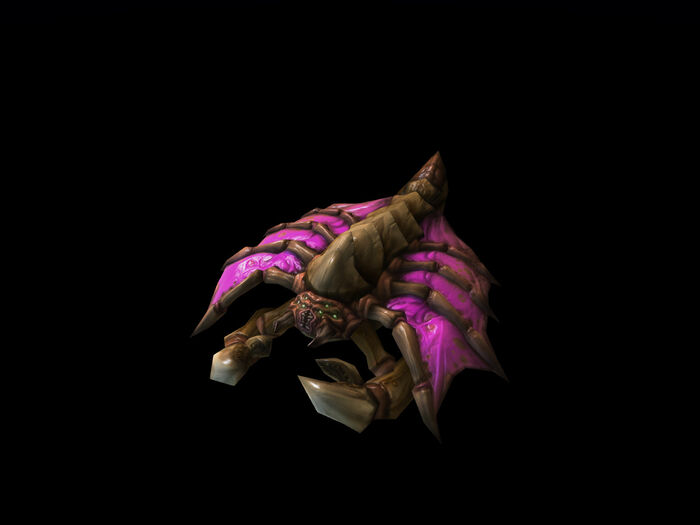}}
  \caption{ The StarCraft units that have been used in this work: (a) marine, (b) zergling, (c) void ray and (d) drone.  }
  \label{f:units}
  \end{center}
\end{figure}

Figure~\ref{f:units} shows the race units have been considered in this work. Table~\ref{t:1} gives the corresponding unit parameters. Interested readers can try other units by editing the map developed in this work with StarCraft map editor. 

\begin{table}[ht]
\caption
	{The information of the units \cite{SCunits} used in this paper. 
	}\label{t:1}
\begin{center}
\begin{tabular}{c c c c c c }
\toprule
\midrule
Name     &	Health $H$	&	Sight range 	 & Attack range $r$  &  Speed $U$    & 	Damage per second DPS \\
\hline
Marine  &  45	& 9  & 5  & 3.15  & 9.8   \\

Zergling  & 35 & 8  &  0.1  & 4.13 & 10   \\

Drone & 40 & 8  &  0.1  & 3.94 & 4.67   \\

Void ray & 150 & 10  &  6  & 3.85 & 16.8   \\
\bottomrule
\end{tabular}
\end{center}
\end{table}

\subsection{Some of the main code subroutines}
\begin{itemize}
\item exec\_2agents.py: the adversary-agent program entry point, will set up the neural network architecture and conduct the fit operation.
\item sc2DqnAgent.py: defines the DQN agent. 
\item agent.py: is to be inherited by sc2DqnAgent.py, and defines the key fit function.   
\item env.py: sets up the StarCraft II environment, and defines the possible actions and the key step function.   
\item sc2\_env\_xun.py: is to be inherited by env.py, and extends the StarCraft II environment to the pursuit--evasion problem.
\item TestScripted\_V1.py: the script tests a traversal algorithm for the pursuers in  
  the FindAndDefeatZerglings mini-game. 
\item TestScripted\_V2.py: the script tests a traversal algorithm for the pursuers in  
  the FindAndDefeatDrones mini-game.   
\end{itemize}
\noindent Other files are from Keras-rl, only after slight modifications (most of them should have been explicitly pointed out in code annotation).

\renewcommand{\thefigure}{B\arabic{figure}}
\setcounter{figure}{0}

\end{document}